\def\preprintversion{1}
\documentclass[twoside,11pt]{article}

\ifdefined\preprintversion\usepackage[preprint]{jmlr2e}\else\usepackage{jmlr2e}\fi

\usepackage[utf8]{inputenc}
\usepackage[T1]{fontenc}
\usepackage{microtype}

\usepackage{amsmath}
\usepackage{mathtools}
\usepackage{bm}
\allowdisplaybreaks

\usepackage{xcolor}
\definecolor{ForestGreen}{RGB}{34,139,34}
\definecolor{Green}{RGB}{0,128,0}
\definecolor{Red}{RGB}{200,0,0}
\definecolor{Top1}{HTML}{0072B2}
\definecolor{Top2}{HTML}{E69F00}
\definecolor{Top3}{HTML}{CC79A7}

\usepackage{booktabs}
\usepackage{multirow}
\usepackage{makecell}
\usepackage{adjustbox}   
\usepackage{pifont}      
\usepackage{etoc}        

\usepackage{tikz}
\usetikzlibrary{arrows.meta, positioning, fit, calc, backgrounds, quotes,
                decorations.pathreplacing, decorations.pathmorphing,
                shapes, shapes.geometric}

\usepackage{listings}
\definecolor{codegray}{RGB}{245,245,245}
\definecolor{codekw}{RGB}{0,60,130}
\definecolor{codestr}{RGB}{130,60,0}
\definecolor{codecmt}{RGB}{90,120,90}
\lstdefinestyle{sheafpython}{
    language=Python,
    basicstyle=\small\ttfamily,
    keywordstyle=\color{codekw}\bfseries,
    stringstyle=\color{codestr},
    commentstyle=\color{codecmt}\itshape,
    backgroundcolor=\color{codegray},
    frame=single,
    framerule=0pt,
    xleftmargin=6pt,
    xrightmargin=6pt,
    aboveskip=6pt,
    belowskip=4pt,
    showstringspaces=false,
    columns=fullflexible,
    keepspaces=true,
}

\renewenvironment{proof}[1][Proof]{\par\noindent{\bf #1\ }}{\hfill\BlackBox\\[2mm]}

\usepackage[most]{tcolorbox}
\tcbuselibrary{skins, breakable}

\definecolor{proofbar}{RGB}{225,225,225}
\definecolor{theorembar}{RGB}{128, 0, 128}
\definecolor{propositionbar}{RGB}{0, 100, 0}
\definecolor{definitionbar}{RGB}{0, 0, 200}
\definecolor{lemmabar}{RGB}{255, 140, 0}
\definecolor{corollarybar}{RGB}{204, 153, 255}
\definecolor{remarkbar}{RGB}{0, 139, 139}
\definecolor{assumptionbar}{RGB}{178, 34, 34}
\definecolor{examplebar}{RGB}{70, 70, 70}

\tcbset{
    mybarstyle/.style={
        enhanced,
        breakable,
        sharp corners,
        colback=white,
        frame hidden,
        boxrule=0pt,
        borderline west={2pt}{0pt}{#1},
        left=6pt,
        right=0pt,
        top=0pt,
        bottom=0pt,
        boxsep=0pt,
        before skip=10pt,
        after skip=10pt,
    }
}

\makeatletter
\newcommand{\barwrapenvironment}[2]{%
  \expandafter\let\csname barwrap@#1\expandafter\endcsname\csname #1\endcsname
  \expandafter\let\csname barwrap@end#1\expandafter\endcsname\csname end#1\endcsname
  \expandafter\renewcommand\csname #1\endcsname{%
    \tcolorbox[mybarstyle=#2]\csname barwrap@#1\endcsname}%
  \expandafter\renewcommand\csname end#1\endcsname{%
    \csname barwrap@end#1\endcsname\endtcolorbox}%
}
\makeatother

\newif\iftheorembarsenabled
\newcommand{\enableappendixtheorembars}{%
  \global\theorembarsenabledtrue
  \barwrapenvironment{theorem}{theorembar}%
  \barwrapenvironment{proposition}{propositionbar}%
  \barwrapenvironment{lemma}{lemmabar}%
  \barwrapenvironment{corollary}{corollarybar}%
  \barwrapenvironment{definition}{definitionbar}%
  \barwrapenvironment{example}{examplebar}%
  \barwrapenvironment{remark}{remarkbar}%
  \barwrapenvironment{assumption}{assumptionbar}%
  \barwrapenvironment{proof}{proofbar}%
}

\newcommand{\restatement}[3]{%
  \iftheorembarsenabled
    \ifcsname restatebar#1\endcsname
      \edef\restatebarcolour{\csname restatebar#1\endcsname}%
    \else
      \def\restatebarcolour{theorembar}%
    \fi
    \begin{tcolorbox}[mybarstyle=\restatebarcolour]%
      \noindent{\bf #1~\ref{#2}}\ \ {\itshape #3}%
    \end{tcolorbox}%
  \else
    \par\medskip\noindent{\bf #1~\ref{#2}}\ \ {\itshape #3}\par\medskip
  \fi}

\providecommand{\pmstd}[2]{%
  \if\relax\detokenize{#1}\relax\textemdash\else#1{\scriptsize$\pm$#2}\fi}
\providecommand{\placeholder}[1]{\textemdash}

\definecolor{ufill}{HTML}{E1D5E7}
\definecolor{uborder}{HTML}{9673A6}
\definecolor{vfill}{HTML}{D5E8D4}
\definecolor{vborder}{HTML}{82B366}

\newcommand{\red}[1]{\textcolor{red}{#1}}

\hypersetup{
   colorlinks=true,
   citecolor=orange,
}

\usepackage{bm}

\newcommand{\R}{\mathbb{R}}

\newcommand{\bT}{\mathbf{T}}
\newcommand{\btheta}{\boldsymbol{\theta}}
\newcommand{\defeq}{\coloneqq}
\DeclareMathOperator{\spec}{spec}

\def\eqref#1{equation~\ref{#1}}

\def\1{\bm{1}}

\def\rvx{{\mathbf{x}}}

\def\rmA{{\mathbf{A}}}

\def\rmD{{\mathbf{D}}}

\def\rmI{{\mathbf{I}}}

\def\rmW{{\mathbf{W}}}
\def\rmX{{\mathbf{X}}}

\DeclareMathAlphabet{\mathsfit}{\encodingdefault}{\sfdefault}{m}{sl}

\def\gE{{\mathcal{E}}}
\def\gF{{\mathcal{F}}}
\def\gG{{\mathcal{G}}}

\def\gV{{\mathcal{V}}}

\def\sR{{\mathbb{R}}}

\DeclareMathOperator*{\argmin}{arg\,min}

\DeclarePairedDelimiter{\parens}{\lparen}{\rparen}

\DeclareMathOperator{\F}{\mathcal{F}}

\DeclareMathOperator{\x}{\mathbf{x}}
\DeclareMathOperator{\X}{\mathbf{X}}

\DeclareMathOperator{\z}{\mathbf{z}}
\DeclareMathOperator{\Z}{\mathbf{Z}}
\DeclareMathOperator{\W}{\mathbf{W}}
\DeclareMathOperator{\I}{\mathbf{I}}

\newcommand{\calT}{\mathcal{T}}
\newcommand{\calR}{\mathcal{R}}

\ifdefined\preprintversion\enableappendixtheorembars\fi

\renewcommand{\eqref}[1]{(\ref{#1})}

\usepackage{lastpage}

\newcommand{\libname}{\textsf{Sheaf-MPNN}}
\newcommand{\liburl}{\url{https://github.com/alessioborgi/sheaf-mpnn}}

\newcommand{\stmtrecovery}{%
The following architectures admit the SheafMPNN representation in Equations~\eqref{eqn:sheafmpnn-transform}--\eqref{eqn:sheafmpnn-update}, with matching normalization and updates:
(i) Neural Sheaf Diffusion;
(ii) standard MPNNs, via identity restrictions and identity pre-transform;
(iii) SheafAN, Res-SheafAN, and sheaf adjacency propagation; and
(iv) Neural Sheaf Propagation, on an augmented sheaf with node stalks $\R^{2d}$ and edge stalks $\R^d$.}

\jmlrheading{27}{2026}{1-\pageref{LastPage}}{8/26}{--}{26-0000}{Alessio Borgi et al.}

\ShortHeadings{\libname{}: Sheaf Neural Networks as Message Passing}{Borgi et al.}
\firstpageno{1}

\begin{document}

\title{\libname{}: A Library for Sheaf Neural Networks\\as Message Passing}

\author{\name Alessio Borgi$^{1,2,\ast}$ \email {[alessio.borgi@uniroma1.it, ab3352@cam.ac.uk]}\\
        \name Mario Severino$^{1}$ \email{[ms3413@cam.ac.uk]}\\
        \name Luke Braithwaite$^{1}$ \email [lb2027@cam.ac.uk]\\
        \name Steve Azzolin$^{3}$ \email {[steve.azzolin@unitn.it]}\\
        \name R{\'e}mi Bourgerie$^{4}$ \email [remibo@kth.se]\\
        \name Olga Zaghen$^{5}$ \email [o.zaghen@uva.nl]\\
        \name Emanuele Mule$^{6}$ \email [emanuele.mule@telecom-paris.fr]\\
        \name Riccardo Ali$^{1}$ \email [rma55@cam.ac.uk]\\
        \name Leo Minh Kustermann$^{7}$ \email [lkustermann@ethz.ch]\\
        \name Edoardo Coppola$^{1}$ \email [ec2013@cam.ac.uk]\\
        \name Antonio Purificato$^{2}$ \email [purificato@diag.uniroma1.it]\\
        \name Stefano Fiorini$^{8}$ \email{[s.fiorini1994@gmail.com]}\\
        \name Christopher Irwin$^{1}$ \email{[cli23@cam.ac.uk]}\\
        \name Filippo Spiga$^{9}$ \email { [fspiga@nvidia.com]}\\
        \name Ling Guo$^{10}$ \email { [lguo@shnu.edu.cn]}\\
        \name Fabrizio Silvestri$^{2}$ \email { [fsilvestri@diag.uniroma1.it]}\\
        \name Michael Bronstein$^{11, 12}$ \email { [michael.bronstein@cs.ox.ac.uk]}\\
        \name Pietro Li\`o$^{1}$ \email { [pl219@cam.ac.uk]}\\
        \addr $^{1}$Department of Computer Science and Technology, University of Cambridge, Cambridge, UK\\
        $^{2}$Department of Computer, Control and Management Engineering, Sapienza University of Rome, Rome, Italy\\
        $^{3}$Department of Information Engineering and Computer Science, University of Trento, Trento, Italy\\
        $^{4}$KTH Royal Institute of Technology, Stockholm, Sweden\\
        $^{5}$AMLab, University of Amsterdam, Amsterdam, Netherlands\\
        $^{6}$Télécom Paris, Institute Polythechnique de Paris, Paris, France\\
        $^{7}$ETH Zürich\\
        $^{8}$Independent researcher\\
        $^{9}$NVIDIA\\
        $^{10}$Department of Mathematics, Shanghai Normal University, Shanghai, China\\
        $^{11}$Department of Computer Science, University of Oxford, Oxford, UK\\
        $^{12}$AITHYRA, Vienna, Austria\\
        $^{\ast}$Corresponding Author.}

\ifdefined\preprintversion\else\editor{\red{[Assigned action editor]}}\fi

\maketitle

\begin{abstract}%
Standard Graph Neural Networks (GNNs) implicitly assume that neighboring nodes share the same feature space, which contributes to oversmoothing and poor performance on heterophilic graphs. Sheaf Neural Networks (SNNs) remove this assumption by giving each node and edge its own vector space and learning restriction maps that transform features as they move between neighbors. These maps define a sheaf Laplacian that drives diffusion over the graph, but existing implementations often assemble it explicitly as a global sparse block matrix, which adds memory and computational overhead and can hinder scalability. In this paper, we introduce \libname{}, a PyTorch Geometric library that reformulates sheaf diffusion as local incidence-wise transport and aggregation, eliminating the need to construct the global Laplacian. \libname{} implements fifteen families of SNNs under a unified message-passing framework, each supporting both homogeneous and heterogeneous graphs, and directed and undirected settings, with diagonal, general, orthogonal, and low-rank restriction maps and attention-based predictors. We further provide a unified benchmarking suite covering node- and graph-level tasks, heterogeneous graph learning, and computational efficiency, enabling systematic and reproducible comparisons across architectures. We establish theoretical equivalence between these message-passing formulations and their original counterparts and identify the architectural mechanisms needed to preserve their propagation dynamics. This unified perspective provides a principled foundation for implementing, comparing, and extending sheaf-based architectures within standard graph learning frameworks. Source code is available at \liburl{}.
\end{abstract}


\section{Introduction}
\label{sec:introduction}
Graph Neural Networks (GNNs) typically learn graph-structured representations through the message-passing paradigm, in which nodes update their representations by aggregating information from their neighbours~\citep{gilmer2017messagepassing}. Sheaf Neural Networks (SNNs) enrich graph learning with \emph{cellular sheaves}, which assign local vector spaces, or \emph{stalks}, to nodes and edges, together with linear \emph{restriction maps} for each node--edge incidence. These maps transport node representations into a shared edge stalk, where neighbouring signals can be compared. SNNs use this structure to define propagation operators, commonly based on the sheaf Laplacian, that allow neighbouring representations to remain consistent after transport even when their original features differ~\citep{hansen2020sheaf,bodnar2022neural}. Subsequent research has explored how sheaves are constructed~\citep{barbero2022connectionLaplacian,caralt2024joint,gillespie2024bayesian}, how information propagates through them~\citep{suk2022surfing,barbero2022attention,zaghen2024nonlinear,bamberger2025bundle,borgi2026polynomialneuralsheafdiffusion,bourgerie2026deep}, and extensions to different graph structures and feature spaces~\citep{braithwaite2026heterogeneous,fiorini2026sheaves,duta2023hypersheaf,mule2026directional,borgi2026esnn,ribeiro2026gaussian, ali2026topa, vanlooy2026onda}. Despite their common foundations, these models differ in their restriction maps, propagation operators, normalization schemes, and update rules. Appendix~\ref{app:primer} surveys these developments.

\noindent A common computational challenge arises from the explicit construction of sheaf operators. Existing SNN implementations assemble the global sheaf Laplacian as a sparse block matrix and apply it to the stacked node representations. For a graph with $n$ nodes and $d$-dimensional stalks, this produces an $nd \times nd$ operator whose nonzero blocks may contain dense $d \times d$ matrices. Constructing and storing this operator introduces additional memory and computational overhead, particularly for large graphs or higher-dimensional stalks. Yet the action of the sheaf Laplacian is inherently local: each edge compares its endpoint representations after transport into a shared stalk, and the resulting contributions are mapped back to the incident nodes and aggregated. This suggests that sheaf propagation can be implemented directly through local message passing, without explicitly assembling the global Laplacian. Such a formulation removes the need for operator assembly, although its practical efficiency depends on the implementation and must be evaluated empirically.

\noindent Motivated by this observation, we introduce \libname{}, a PyTorch Geometric library that implements sheaf propagation through local incidence-wise transport and aggregation. By expressing different SNN architectures within a common computational interface, \libname{} facilitates their implementation, reuse, and comparison while integrating them with established graph learning tools. The library provides 15 families of SNNs, each with homogeneous and heterogeneous variants, and supports directed and undirected graph settings. It includes diagonal, general, orthogonal, and low-rank restriction maps, together with attention-based map predictors where applicable. To enable systematic evaluation, we also provide a unified benchmarking suite, covering more than 20 datasets, including node- and graph-level learning, heterogeneous graph benchmarks, and computational efficiency, with shared experimental configurations and reproducible evaluation protocols.

\noindent Beyond the library itself, we also develop the unified Sheaf Message Passing Neural Network (Sheaf-MPNN) theoretical framework that characterises how existing sheaf architectures can be expressed through local message passing. We derive explicit message-passing formulations for 15 architecture configurations and establish their theoretical equivalence to the original models, identifying the mechanisms required to preserve their propagation dynamics. These include augmented node states, type-conditioned restriction maps, and directed aggregation rules. We further show that the \textsc{SheafPool} graph-level readout ~\citep{braithwaite2026heterogeneous} can be formulated within the same framework by aggregating locally aligned node representations into a receive-only virtual node. Together, our contribution provides a common theoretical and computational foundation for implementing, comparing, and extending sheaf-based graph learning. Appendix~\ref{app:library} details the supported architectures, implementation choices, and normalization conventions.

\section{Sheaf Message Passing in a Nutshell}
\label{sec:blueprint}
Let $\gG=(\gV,\gE)$ be a finite undirected graph without self-loops, and let $\gF$ be a cellular sheaf assigning finite-dimensional inner-product spaces $\gF(u)$ and $\gF(e)$ to nodes and edges, with linear restriction maps $\F_{u\trianglelefteq e}:\gF(u)\to\gF(e)$ for each incidence $u\trianglelefteq e$~\citep{curry2014sheaves,hansen2020sheaf}. Node signals belong to $C^0(\gG,\gF)=\bigoplus_{u\in\gV}\gF(u)$. In orthonormal stalk bases, the sheaf Laplacian acts on $\x\in C^0(\gG,\gF)$ as:
\begin{equation}
 (L_{\gF}\x)_u=\sum_{e=\{u,v\}\in\gE}
 \F_{u\trianglelefteq e}^{\top}
 \left(\F_{u\trianglelefteq e}\x_u-\F_{v\trianglelefteq e}\x_v\right).
 \label{eq:local-laplacian}
\end{equation}
Here, each term measures disagreement in the shared edge stalk and transports it back to the receiving node, revealing the Laplacian as a local aggregation operator. We generalise this observation into a \emph{Sheaf Message Passing Neural Network (Sheaf-MPNN)}. Assuming stalks of dimension $d$ and $f$ feature channels, each node has state $\x_u^{(k)}\in\R^{d\times f}$ at layer $k$. A SheafMPNN transforms node states, computes interactions in edge stalks, aggregates the resulting messages, and updates node representations as:
\begin{align}
    \z_u^{(k)} &= \W_1^{(k)}\x^{(k)}_u\W_2^{(k)},
        \label{eqn:sheafmpnn-transform}\\
    \mathbf{m}^{(k)}_{u \trianglelefteq e} &= \psi_e\parens*{\F_{v \trianglelefteq e}\z_v^{(k)}, \F_{u \trianglelefteq e}\z_u^{(k)}, \mathbf{e}_{uv}},
        \label{eqn:sheafmpnn-message}\\
    \mathbf{m}^{(k+1)}_u &= \bigoplus_{e=\{u,v\}\in\gE} \omega_{uv}^{(k)} \F_{u \trianglelefteq e}^{\top} \mathbf{m}^{(k)}_{u \trianglelefteq e},
        \label{eqn:sheafmpnn-aggregate}\\
    \x_{u}^{(k+1)} &= \varphi_u\parens*{\x_{u}^{(k)}, \mathbf{m}^{(k+1)}_u}.
        \label{eqn:sheafmpnn-update}
\end{align}
Here, $\W_1^{(k)}\in\R^{d\times d}$ and $\W_2^{(k)}\in\R^{f\times f}$ mix stalk coordinates and feature channels, respectively. The function $\psi_e$ computes interactions between transported representations, optionally incorporating edge features $\mathbf{e}_{uv}$. Each message is weighted by $\omega_{uv}^{(k)}$, pulled back through $\F_{u\trianglelefteq e}^{\top}$, aggregated by $\bigoplus$, and combined with the current state through $\varphi_u$. Restriction maps may be learned at each layer, with their layer index omitted for clarity. The construction is equivariant to node relabelling when aggregation is permutation invariant, and the map predictors and update functions are shared and independent of node ordering. Figure~\ref{fig:sheaf-mpnn} illustrates the distinction from standard message passing. Sheaf diffusion is recovered by setting $\psi_e(a,b)=b-a$, $\omega_{uv}^{(k)}=1$, and using sum aggregation, yielding Equation~\eqref{eq:local-laplacian} on the transformed features. Normalization is incorporated through $\F_{u\trianglelefteq e}N_u$ and its transpose, where the implementation uses $N_u=(D_u+I)^{-1/2}$ with $D_u=\sum_{e\ni u}\F_{u\trianglelefteq e}^{\top}\F_{u\trianglelefteq e}$ (Appendix~\ref{app:library-layer}). Conversely, identity restrictions and identity pre-transformations recover standard MPNNs. The framework also accommodates heterogeneous graphs through type-conditioned restriction-map predictors (Appendix~\ref{app:hetsheaf}) and directed graphs through incoming-arc aggregation (Appendix~\ref{app:directed}). Explicit directed propagation is necessary because an undirected edge yields cross-edge blocks $\F_{u\trianglelefteq e}^{\top}\F_{v\trianglelefteq e}$ and $\F_{v\trianglelefteq e}^{\top}\F_{u\trianglelefteq e}$ that are transposes, rather than independent directional operators. We formalise the recovery of existing architectures below. 
\begin{figure}[t]
    \centering
    \scalebox{0.7}{\maxsizebox{\textwidth}{!}{%
\begin{tikzpicture}[
    every edge quotes/.style={font=\footnotesize},
]
    \begin{scope}[local bounding box=mpnn]
        \node[circle, fill=red!20!white] (x0) {$\mathbf{x}_0$};

        \node[circle, above left=of x0, fill=green!20!white] (m01) {$m_{01}$};
        \node[circle, above right=of x0, fill=green!20!white] (m02) {$m_{02}$};
        \node[circle, below right=of x0, fill=green!20!white] (m03) {$m_{03}$};
        \node[circle, below left=of x0, fill=green!20!white] (m04) {$m_{04}$};

        \node[circle, above left=of m01, fill=blue!20!white] (x1) {$\mathbf{x}_1$};
        \node[circle, above right=of m02, fill=blue!20!white] (x2) {$\mathbf{x}_2$};
        \node[circle, below right=of m03, fill=blue!20!white] (x3) {$\mathbf{x}_3$};
        \node[circle, below left=of m04, fill=blue!20!white] (x4) {$\mathbf{x}_4$};

        \path[-stealth, dashed, gray, thick]
            (x0) edge[bend right] (m01)
            (x0) edge[bend left] (m02)
            (x0) edge[bend right] (m03)
            (x0) edge[bend left] (m04)
            (x1) edge[bend right] (m01)
            (x2) edge[bend left] (m02)
            (x3) edge[bend right] (m03)
            (x4) edge[bend left] (m04);

        \path[-stealth, ultra thick]
            (m01) edge (x0)
            (m02) edge (x0)
            (m03) edge (x0)
            (m04) edge (x0);
    \end{scope}

    \begin{scope}[
        local bounding box=sheaf-mpnn,
        shift={($(mpnn.east)+(8cm,0)$)}
    ]
        \node[circle, fill=red!20!white] (y0) {$\mathbf{x}_0$};

        \node[above left=of y0, draw] (F01) {$\mathcal{F}(e_{01})$};
        \node[above right=of y0, draw] (F02) {$\mathcal{F}(e_{02})$};
        \node[below right=of y0, draw] (F03) {$\mathcal{F}(e_{03})$};
        \node[below left=of y0, draw] (F04) {$\mathcal{F}(e_{04})$};

        \node[circle, above left=of F01, fill=blue!20!white] (y1) {$\mathbf{x}_1$};
        \node[circle, above right=of F02, fill=blue!20!white] (y2) {$\mathbf{x}_2$};
        \node[circle, below right=of F03, fill=blue!20!white] (y3) {$\mathbf{x}_3$};
        \node[circle, below left=of F04, fill=blue!20!white] (y4) {$\mathbf{x}_4$};

        \path[-stealth, dashed, gray, thick]
            (y0) edge[bend right] (F01)
            (y0) edge[bend left] (F02)
            (y0) edge[bend right] (F03)
            (y0) edge[bend left] (F04)
            (y1) edge[bend right] (F01)
            (y2) edge[bend left] (F02)
            (y3) edge[bend right] (F03)
            (y4) edge[
                bend left,
                "$\F_{4\trianglelefteq e_{04}}\x_4$"'{below right}
            ] (F04);

        \node[
            font=\footnotesize,
            fill=white,
            inner sep=0pt
        ] at ($(F02.north)+(-25pt,10pt)$) (msg)
        {$\psi\parens*{
            \F_{2\trianglelefteq e_{02}}\x_2,
            \F_{0\trianglelefteq e_{02}}\x_0
        }$};

        \path[-stealth, thick]
            (msg) edge[bend right] (F02);

        \path[-stealth, ultra thick]
            (F01) edge (y0)
            (F02) edge[
                "$\F_{0\trianglelefteq e_{02}}^{\top}$"' below right
            ] (y0)
            (F03) edge (y0)
            (F04) edge (y0);
    \end{scope}

    \node[
        fit=(sheaf-mpnn),
        inner sep=10pt,
        rounded corners,
        label={below:Sheaf MPNN}
    ] {};

    \node[
        fit=(mpnn),
        inner sep=10pt,
        rounded corners,
        label={below:MPNN}
    ] {};
\end{tikzpicture}%
}}
    \caption{\textbf{Sheaf Message Passing Neural Network (SheafMPNN).}
    Neighbourhood update at node $0$ (red), with neighbouring nodes in blue.
    \textbf{(Left)} A standard MPNN computes messages $m_{0i}$ (green) from neighbouring node states and aggregates them at node $0$.
    \textbf{(Right)} A SheafMPNN transports both endpoint representations into the shared edge stalk $\mathcal{F}(e_{0i})$ (boxes) through restriction maps (dashed grey arrows). The function $\psi$ computes their interaction, and the resulting message is pulled back through $\F_{0\trianglelefteq e_{0i}}^{\top}$ and aggregated at node $0$ (solid black arrows).}
    \label{fig:sheaf-mpnn}
    \vspace{-0.75cm}
\end{figure}
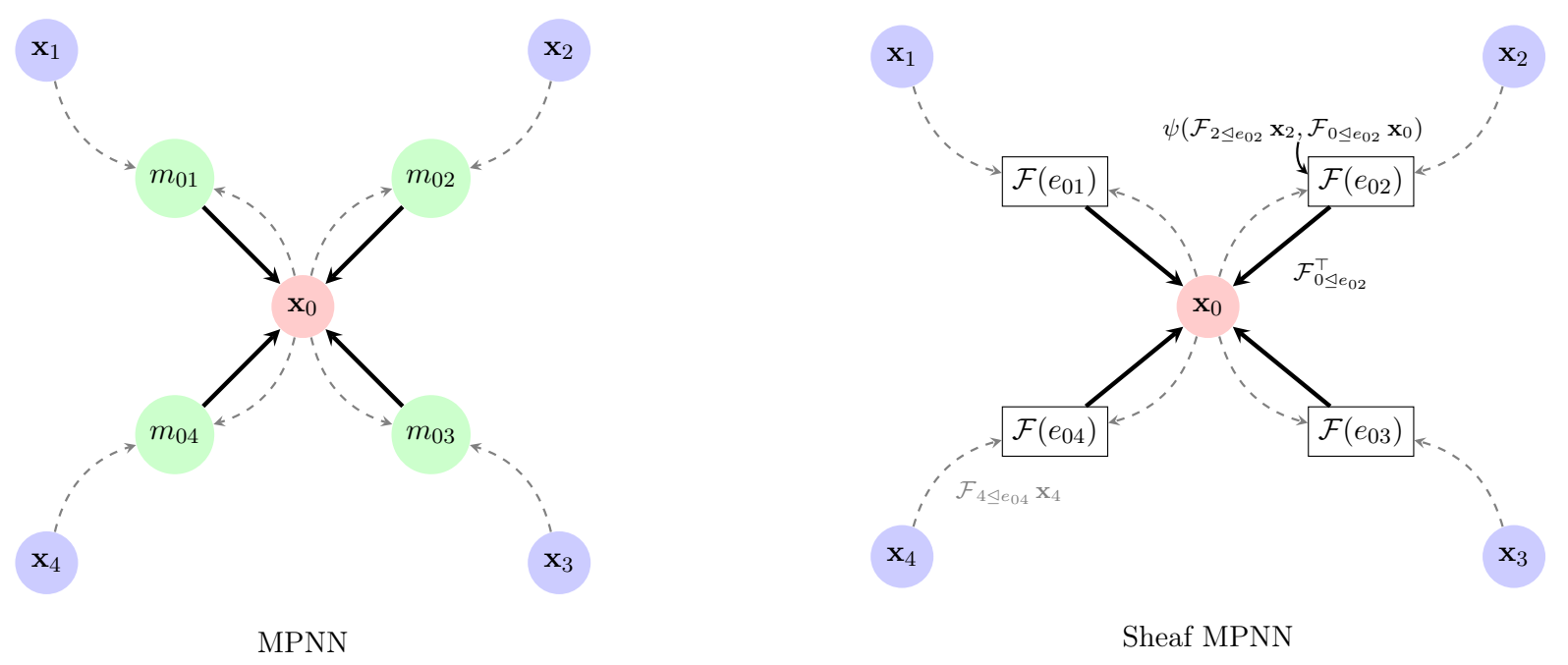
\begin{theorem}
\label{thm:recovery}
\stmtrecovery
\end{theorem}
\noindent Appendix~\ref{app:equivalence} proves Theorem~\ref{thm:recovery} and extends the framework to diagonal, general, orthogonal, low-rank, and attention-generated NSD restrictions, as well as to all the sheaf-variants (Theorems ~\ref{thm:polynsd}, ~\ref{prop:sheafpool}, ~\ref{thm:hetnsd}, ~\ref{thm:hetnsp}, and ~\ref{thm:dirsheafmpnn}--\ref{thm:bayesian}). 

\noindent The computational cost of SheafMPNN reflects the locality of its propagation operator: the transport and aggregation require $O(|\gE|d^2f)$ operations, with $O(|\gE|d^2+|\gE|df+ndf)$ memory for restriction maps, edge messages, and node states when messages are materialised. These bounds exclude map prediction, channel mixing, and block normalization. So, at the end, we evaluate the sheaf Laplacian entirely through local operations, without assembling its global $nd\times nd$ matrix representation (Remark~\ref{rem:laplacian_free}).

\section{Library Design and Usage}
\label{sec:design}

Built on PyTorch~\citep{paszke2019pytorch} and PyTorch Geometric~\citep{fey2025pyg}, \libname{} provides reusable sheaf message-passing layers and end-to-end model wrappers through a unified interface (Appendix~\ref{app:library-families}). The following example illustrates node classification using orthogonal Neural Sheaf Diffusion:

\begin{lstlisting}[style=sheafpython]
from torch_geometric.datasets import HeterophilousGraphDataset
from sheaf_mpnn.nsd import NSDModel, NSDVariant

data = HeterophilousGraphDataset("data", name="Roman-empire")[0]
model = NSDModel(in_channels=data.num_features, out_channels=18,
                 hidden_dim=16, num_layers=4, stalk_dim=4,
                 variant=NSDVariant.ORTHOGONAL)
out = model(data.x, data.edge_index)  # [num_nodes, 18]
\end{lstlisting}

\noindent The accompanying \texttt{sheaf} command-line interface supports reproducible experiments across more than 20 datasets, Optuna-based hyperparameter optimization~\citep{optuna_2019}, and computational efficiency comparisons against reference implementations. Appendix~\ref{app:library} details the package structure, model interfaces, and experimental tooling.

\section{Related Software}
\label{sec:related}

General-purpose frameworks such as PyTorch Geometric~\citep{fey2025pyg} and DGL~\citep{wang2019dgl} provide message-passing abstractions without native support for cellular sheaves. TopoX~\citep{hajij2024topox} supports higher-order topological domains, including operator-based sheaf models, while existing SNN implementations~\citep{bodnar2022neural,barbero2022attention,braithwaite2026heterogeneous} primarily target individual architectures and often rely on explicit sheaf-Laplacian assembly. In contrast, \libname{} provides a reusable, Laplacian-free message-passing implementation across a broad family of SNNs, complemented by unified benchmarking tools and theoretical equivalence results.

\ifdefined\preprintversion\else
\section{Benchmarks}
\label{sec:benchmarks}

We evaluate \libname{} in terms of predictive performance and computational efficiency. Table~\ref{tab:new_heterophily_results} reports results on the heterophily benchmarks of \citet{platonov2023critical}, averaged over 10 fixed splits, alongside published results from the corresponding Laplacian-based architectures. Appendix~\ref{app:experiments} extends this evaluation to the benchmarks of \citet{pei2020geom} and the supported restriction-map families. To assess computational efficiency, Table~\ref{tab:efficiency} compares peak GPU memory and per-epoch runtime against the original NSD implementation under matched architectural configurations and hyperparameters, including a synthetic scalability study across increasing graph sizes.

\red{[Insert headline accuracy, memory, runtime, and scalability results once experiments are complete.]}

\input{tables/heterophily_main}
\input{tables/efficiency}

\fi

\ifdefined\preprintversion
\section{Availability}
\else
\section{Availability and Roadmap}
\fi
\label{sec:availability}

\libname{} is publicly available at \liburl{} under the CC-BY 4.0 license, together with documentation and tutorial notebooks. Appendix~\ref{app:library} provides implementation details, supported architectures, and experimental configurations.

\ifdefined\preprintversion\else
Planned extensions include support for hypergraph sheaves and sheaf-based positional encodings.
\fi

\ifdefined\preprintversion\else
\acks{\red{[Acknowledgements. JMLR requires a funding disclosure (financial activities supporting the submitted work) and competing-interests statement (related financial activities outside the submitted work).]}
The computations were partly enabled by resources provided by the National Academic Infrastructure for Supercomputing in Sweden (NAISS), partially funded by the Swedish Research Council through grant agreement no. 2022-06725.}

\fi
\newpage
\appendix
\ifdefined\preprintversion\else\enableappendixtheorembars\fi

\begin{center}
    {\LARGE \bfseries \libname{}: A Library for Sheaf Neural Networks as Message Passing \par}
    \vspace{0.4em}
    {\Large Supplementary Material\par}
\end{center}

\vspace{1em}

\addcontentsline{toc}{part}{Supplementary Material}
\etocsetnexttocdepth{3}
\vspace{-1cm}
\localtableofcontents

\vspace{1em}

\newpage
\section{Primer on Cellular Sheaves and Sheaf Neural Networks}
\label{app:primer}

This appendix introduces in a deeper way the Sheaf Neural Networks (SNNs) and places the  \libname{} library within the literature. It first explains how a cellular sheaf generalizes the graph Laplacian and why diffusion on a sheaf avoids the classical failure modes of message passing, and then reviews how the field has developed. For fuller treatments we refer to \citet{curry2014sheaves} and \citet{rosiak2022sheaf} for cellular sheaf theory, to \citet{hansen2019toward} for the spectral theory of the sheaf Laplacian, and to the survey of \citet{ayzenberg2025sheaf} for sheaf theory in machine learning more broadly.

\subsection{From Message Passing to Sheaves}
\label{app:primer-mpnn}

\paragraph{Message Passing Neural Networks.} A message passing neural network~\citep{gilmer2017messagepassing} updates the representation $\rvx_u$ of each node $u$ of a graph $\gG=(\gV,\gE)$ by aggregating the states of its neighbours $\mathcal{N}(u)$:
\begin{equation}
    \rvx_u^{(t+1)} = \mathrm{Update}^{(t)}\!\left(\rvx_u^{(t)},\; \mathrm{Aggregate}^{(t)}\!\left(\left\{ \rvx_v^{(t)} : v \in \mathcal{N}(u) \right\}\right)\right).
    \label{eq:primer-mpnn}
\end{equation}
This update assumes that all node features live in one shared vector space, so that neighbouring features can be compared and averaged directly. The assumption is harmless when connected nodes tend to be similar, but it underlies two well-known failures. Repeated low-pass averaging drives features towards a constant signal as depth grows, a phenomenon known as oversmoothing~\citep{cai2020noteOversmoothing, rusch2023survey}; and on heterophilic graphs, where neighbours tend to belong to different classes, averaging erases the very signal that separates them. Cellular sheaves remove the assumption at its source: every node and every edge receives its own vector space, and learnable linear maps decide how signals are compared across each edge.

\paragraph{Cellular Sheaves.} A sheaf attaches data to the pieces of a space, together with rules stating how the data on overlapping pieces must agree. It therefore describes when local information can be glued into a consistent global picture. Cellular sheaves make this idea discrete by attaching vector spaces to the cells of a cell complex and linear maps to the relations between them. On a graph, the cells are the vertices and the edges, and each incidence $u \trianglelefteq e$, read as ``$u$ is a face of $e$'', carries a linear map that moves data from the vertex to the edge where it is compared with data from the other endpoint.

\begin{definition}[Cellular Sheaf on a Graph]
\label{def:primer-cellular-sheaf}
A cellular sheaf~\citep{shepard1985cellular, curry2014sheaves} on an undirected graph $\gG = (\gV, \gE)$ assigns a finite-dimensional inner-product space (the \emph{stalk}) $\gF(u)$ to each vertex $u \in \gV$ and $\gF(e)$ to each edge $e \in \gE$, together with linear \emph{restriction maps} $\gF_{u \trianglelefteq e}: \gF(u) \to \gF(e)$ for each incident node--edge pair $u \trianglelefteq e$. Stacking vertex and edge stalks yields the $0$- and $1$-cochain spaces $C^0(\gG, \gF) \coloneqq \bigoplus_{u \in \gV} \gF(u)$ and $C^1(\gG, \gF) \coloneqq \bigoplus_{e \in \gE} \gF(e)$, respectively.
\end{definition}
\noindent An intuitive reading comes from opinion dynamics~\citep{hansen2021opinion}. Each node stalk is a private opinion space, each edge stalk is a shared space for the conversation along that edge, and $\gF_{u \trianglelefteq e}$ describes how $u$ expresses its opinion in that conversation. This means that two neighbours, instead of agreeing when their raw features do, they agree along $e$ when their expressed opinions coincide, $\gF_{u \trianglelefteq e}\rvx_u = \gF_{v \trianglelefteq e}\rvx_v$,  Disagreement is therefore measured only after both signals have been aligned in a common space, which is exactly what a heterophilic edge requires.

Orienting each edge as $e = (u \to v)$, the \emph{coboundary} is $\delta: C^0(\gG, \gF) \to C^1(\gG, \gF)$, and the $\delta(\rvx)_e \coloneqq \gF_{v \trianglelefteq e}\rvx_v - \gF_{u \trianglelefteq e}\rvx_u$ records this disagreement edge by edge. Resulting from this, the $0$-cochain with $\delta(\rvx) = 0$ is a \emph{global section}: a configuration that is consistent across the whole graph. The \emph{sheaf Laplacian} $L_{\gF} \coloneqq \delta^{\top}\delta$~\citep{hansen2019toward} acts node-wise as:
\begin{equation}
    L_{\gF}(\rvx)_u = \sum_{u,v \trianglelefteq e} \gF_{u \trianglelefteq e}^{\top}\!\left(\gF_{u \trianglelefteq e}\rvx_u - \gF_{v \trianglelefteq e}\rvx_v\right).
    \label{eq:primer-sheaf-laplacian}
\end{equation}
It is symmetric positive semi-definite, with diagonal blocks $\sum_{u \trianglelefteq e} \gF_{u \trianglelefteq e}^{\top} \gF_{u \trianglelefteq e}$ and off-diagonal blocks $-\gF_{u \trianglelefteq e}^{\top} \gF_{v \trianglelefteq e}$, and its kernel is exactly the space of global sections. Its normalized form is $\Delta_{\gF} \coloneqq D^{-1/2} L_{\gF} D^{-1/2}$, where $D$ is the block diagonal of $L_{\gF}$. We take all stalks to be $\sR^d$, so that $L_{\gF} \in \sR^{nd \times nd}$ for a graph with $n$ nodes. When every restriction map is the identity, Equation~\eqref{eq:primer-sheaf-laplacian} reduces to the graph Laplacian $\sum_{v \in \mathcal{N}(u)} (\rvx_u - \rvx_v)$ and the global sections are the constant signals. Classical graph learning is thus the special case of the \emph{constant sheaf}, which is how the blueprint of Section~\ref{sec:blueprint} recovers standard MPNNs (Theorem~\ref{thm:recovery}).

\paragraph{Sheaf Diffusion and Neural Sheaf Diffusion.}
\label{app:primer-diffusion}
The sheaf Laplacian generates the diffusion $\dot{\rvx}(t) = -\Delta_{\gF}\,\rvx(t)$, which decreases the sheaf Dirichlet energy $\rvx^{\top} L_{\gF}\, \rvx = \|\delta \rvx\|^2$ and converges to the projection of the initial signal onto the global sections. For the constant sheaf these are the constant signals, so deep diffusion oversmooths. For a general sheaf they are shaped by the restriction maps, and may assign different features to neighbouring nodes as long as these agree after transport. \citet{bodnar2022neural} turned this observation into Neural Sheaf Diffusion (NSD). They proved that suitable sheaves let diffusion linearly separate classes in settings, including heterophilic ones, where graph diffusion provably cannot, and they made the sheaf learnable. NSD stacks the stalk signals into $\rmX \in \sR^{nd \times f}$ with $f$ feature channels, predicts the map of every incidence from the features of its two endpoints, $\gF_{u \trianglelefteq e}^{(t)} = \mathbf{\Phi}^{(t)}(\rvx_u \,\|\, \rvx_v)$, as a diagonal, orthogonal, or general matrix, and performs the step:
\begin{equation}
    \rmX^{(t+1)} = \rmX^{(t)} - \sigma\!\left(\Delta_{\gF^{(t)}}\left(\rmI_{n} \otimes \rmW_1^{(t)}\right)\rmX^{(t)}\rmW_2^{(t)}\right),
    \label{eq:primer-nsd}
\end{equation}
where $\rmW_1^{(t)}$ and $\rmW_2^{(t)}$ act on the stalk and channel dimensions, respectively. This is the update that \libname{} evaluates without materializing $\Delta_{\gF^{(t)}}$ (Theorem~\ref{thm:recovery} and Appendix~\ref{app:library-layer}).

\subsection{Sheaf Neural Network Literature}
\label{app:recent-literature}

Sheaf-based learning grew out of applied topology. \citet{hansen2019toward} developed the spectral theory of the sheaf Laplacian, \citet{hansen2019laplacian} showed how to learn it from smooth signals, and \citet{hansen2021opinion} interpreted sheaf diffusion as opinion dynamics. \citet{hansen2020sheaf} then proposed the first sheaf neural networks, with hand-crafted sheaves, and \citet{gebhart2023knowledge} recast knowledge-graph embedding as learning an approximate global section of a \emph{knowledge sheaf}, an idea later used to embed new entities by harmonic extension~\citep{cobb2023feature}. NSD~\citep{bodnar2022neural} made the sheaf itself learnable and supplied the theory on which most subsequent work builds.

A first line of work asks how the sheaf should be obtained. Connection-Laplacian networks~\citep{barbero2022connectionLaplacian} compute orthogonal maps in preprocessing by aligning local tangent spaces, and tangent-bundle networks~\citep{battiloro2024bundle} show that discretizing convolution on a manifold's tangent bundle yields a principled sheaf model. Signal-processing methods learn restriction maps, or entire connection graphs, directly from data~\citep{dinino2024optimizing, dinino2026structured}. Bayesian SNNs~\citep{gillespie2024bayesian} place a variational distribution over sheaves, PAC-Bayes bounds guide the spectral optimization of the learned operator~\citep{choi2026pac}, and eigenvectors of a learned sheaf Laplacian serve as positional encodings for ordinary GNNs~\citep{he2023positionalEncoding}. To interpret what a learned sheaf encodes, the Bochner--Weitzenb\"ock decomposition of \citet{guadagnuolo2026bochner} splits the sheaf Laplacian into a positive semi-definite part and a block-diagonal curvature term, which describes the learned geometry locally rather than through the global spectrum. However, two recent studies temper the enthusiasm: \citet{caralt2026on} find that the fixed identity sheaf is competitive on the standard heterophily benchmarks, and the inductive benchmark of \citet{fiorini2026benchmarking} finds that the choice of restriction maps matters most, yet that SNNs do not surpass the strongest baselines. Both suggest that the value of a learned sheaf depends strongly on the task.

A second line changes how signals move on the sheaf. Continous NSD models the dynamics of Eq.~\eqref{eq:primer-nsd} as continuous time diffusion~\citep{bodnar2022neural}, Sheaf attention networks weight the transported neighbours~\citep{barbero2022attention}, Neural Sheaf Propagation adopts a second-order recurrence~\citep{suk2022surfing}, and nonlinear~\citep{zaghen2024nonlinear} and joint~\citep{caralt2024joint} diffusion modify the interaction law itself. Polynomial filters of the sheaf Laplacian enlarge the receptive field of a single layer~\citep{borgi2026polynomialneuralsheafdiffusion}, while Deep NSD replaces the Laplacian with a sheaf adjacency operator, complemented with per-stalk gating, odd activation and layer normalization, to keep signals informative across many layers~\citep{bourgerie2026deep}. Long-range communication has become a central theme. Cooperative SNNs let each node decide independently whether to send or receive information~\citep{ribeiro2026cooperative}; SheafDEQ formulate sheaf propagation as an equilibrium with convergence guarantees~\citep{bourgerie2026fixed}; oscillatory dynamics over sheaves propagate information through second-order, wave-like updates~\citep{vanlooy2026onda}; and sheaf effective resistance shows that over-squashing can be reduced by changing the sheaf rather than rewiring the graph~\citep{guimaraes2026effective}. Bundle neural networks~\citep{bamberger2025bundle} restrict the sheaf to flat vector bundles and replace message passing with heat-kernel diffusion, and topological attention~\citep{ali2026topa} reinterprets multi-head attention as transport through quiver representations, whose off-diagonal routes mix heads before aggregation.

Sheaves also offer a natural place to encode richer relational structure. Heterogeneous SNNs condition the restriction maps on node and edge types and introduce the gauge-invariant \textsc{SheafPool} readout for graph classification~\citep{braithwaite2026heterogeneous}. Directed SNNs generalize the cellular sheaf to account for edge orientation~\citep{fiorini2026sheaves}, temporal SNNs give each node a time-varying orthogonal frame for link prediction on dynamic graphs~\citep{asif2026temporal}, and hierarchical sheaf pooling coarsens graphs using low-frequency modes of cluster-internal sheaf Laplacians~\citep{naddeo2026hierarchical}. Further work leaves graphs or Euclidean stalks behind. Sheaf hypergraph networks~\citep{duta2023hyper} extend incidence-based transport to group interactions; follow-ups construct hypergraph sheaf Laplacians from symmetric simplicial lifts~\citep{choi2025hypergraph}, add directed hyperedges~\citep{mule2026directional}, and bring sheaves to higher-order message passing in topological deep learning~\citep{hume2025sheafification}. Copresheaf networks~\citep{hajij2025copresheaf} place sheaf models within a broader framework that also covers other deep learning architectures. Other models change the stalks themselves, carrying Gaussian distributions~\citep{ribeiro2026gaussian}, symmetric positive-definite matrices with manifold-compatible operators~\citep{peng2026spd}, Clifford-algebra multivectors~\citep{kamiya2026clifford}, or Euclidean-equivariant vector features~\citep{borgi2026esnn}. In each case the restriction maps, and not only the aggregation, must respect the structure of the stalks.


\section{The SheafMPNN Blueprint Recovers Existing Sheaf Architectures}
\label{app:equivalence}

This appendix proves Theorem~\ref{thm:recovery}: each architecture listed there is an instance of the blueprint in Equations~\eqref{eqn:sheafmpnn-transform}--\eqref{eqn:sheafmpnn-update}, so its layers can be evaluated by local message passing without materializing the sheaf Laplacian. It then extends the construction to polynomial, graph-level, heterogeneous, directed, cooperative, joint, bundle, and Bayesian models. 

\noindent Throughout, $\sigma$ denotes a pointwise nonlinearity, $\rmD_u = \sum_{u \trianglelefteq e}\F_{u \trianglelefteq e}^{\top}\F_{u \trianglelefteq e}$ the block degree of node $u$, and $\Delta_{\gF} = \rmD^{-1/2}L_{\gF}\rmD^{-1/2}$ the normalized sheaf Laplacian. As in Section~\ref{sec:blueprint}, normalization is absorbed into the restriction maps by writing $\hat{\F}_{u \trianglelefteq e} \coloneqq \F_{u \trianglelefteq e}\rmD_u^{-1/2}$, so that:
\begin{equation}
  [\Delta_{\gF}\z]_u = \sum_{u,v \trianglelefteq e} \hat{\F}_{u \trianglelefteq e}^{\top}\bigl(\hat{\F}_{u \trianglelefteq e}\z_u - \hat{\F}_{v \trianglelefteq e}\z_v\bigr).
  \label{eq:normalized-local-laplacian}
\end{equation}
The release instead normalizes with $\mathbf{N}_u = (\rmD_u+\I_d)^{-1/2}$ (Appendix~\ref{app:library-layer}). Since $\rmD_u+\I_d$ is positive definite, this needs no invertible restriction map, and every construction below holds verbatim with $\mathbf{N}_u$ in place of $\rmD_u^{-1/2}$.

\restatement{Theorem}{thm:recovery}{\stmtrecovery}
\noindent In the following, we proof each part of Theorem~\ref{thm:recovery} in turn, showing that the layer update of each architecture can be expressed in the form of Equation~\eqref{eqn:sheafmpnn-update} with a suitable choice of message $\psi_e$, aggregation $\bigoplus$, and update $\varphi_u$.

\begin{proof}[Proof of Theorem~\ref{thm:recovery}, Parts (i)--(ii): Neural Sheaf Diffusion.]
A layer of Neural Sheaf Diffusion (NSD)~\citep{bodnar2022neural} computes:
\begin{align}
  \Z^{(k)} &= \parens*{\I_{n} \otimes \W_1^{(k)}}\X^{(k)}\W_2^{(k)}, \nonumber\\
  \X^{(k+1)} &= (1+\pmb{\varepsilon}^{(k)}) \odot \X^{(k)} - \alpha\,\sigma\parens*{\Delta_{\gF^{(k)}}\Z^{(k)}}
  \label{eqn:nsd-layer}
\end{align}
where $\pmb{\varepsilon}^{(k)} \in \R^{d}$ rescales each stalk dimension, broadcast across nodes and channels, and $\alpha$ is the diffusion step size. By Equation~\eqref{eq:normalized-local-laplacian}, the update at node $u$ is:
\begin{equation*}
  \x_u^{(k+1)} = (1+\pmb{\varepsilon}^{(k)}) \odot \x_u^{(k)} - \alpha\,\sigma\Bigl(\sum_{u,v \trianglelefteq e} \hat{\F}_{u \trianglelefteq e}^{\top}\bigl(\hat{\F}_{u \trianglelefteq e}\z_u^{(k)} - \hat{\F}_{v \trianglelefteq e}\z_v^{(k)}\bigr)\Bigr)
\end{equation*}
with $\z_u^{(k)} = \W_1^{(k)}\x_u^{(k)}\W_2^{(k)}$. This is the blueprint with the normalized maps $\hat{\F}$, the message function $\psi_e(a,b) = b - a$ in Equation~\eqref{eqn:sheafmpnn-message}, unit weights and sum aggregation in Equation~\eqref{eqn:sheafmpnn-aggregate}, and the update $\varphi_u(\x_u, \mathbf{m}_u) = (1+\pmb{\varepsilon}^{(k)}) \odot \x_u - \alpha\,\sigma(\mathbf{m}_u)$ in Equation~\eqref{eqn:sheafmpnn-update}.

For part (ii), take every restriction map and both weight matrices to be identities. The transported states are then the raw node states, the message $\psi_e(\x_v, \x_u, \mathbf{e}_{uv})$ and the aggregation $\bigoplus$ are arbitrary, and the pullback is the identity, so the blueprint reduces to the general MPNN of \citet{gilmer2017messagepassing}.
\end{proof}

\begin{proof}[Proof of Theorem~\ref{thm:recovery}, part (iii): Sheaf Attention.]
Let $\rmA_{\gF}$ be the block matrix whose $(u,v)$ block is $\F_{u \trianglelefteq e}^{\top}\F_{v \trianglelefteq e}$ for every edge $e = \{u,v\}$ and zero otherwise, and let $\hat{\Lambda}^{(k)}$ collect the attention coefficients $\Lambda_{uv} = \mathrm{Attn}(\x_u^{(k)}, \x_v^{(k)})$, normalized over the neighbours of $u$. A SheafAN layer~\citep{barbero2022attention} computes:
\begin{equation}
  \X^{(k+1)} = \sigma\parens*{\parens*{(\hat{\Lambda}^{(k)} \otimes \mathbf{1}_d\mathbf{1}_d^{\top}) \odot \rmA_{\gF^{(k)}}}\Z^{(k)}}
  \label{eqn:att-layer}
\end{equation}
which at node $u$ reads $\x_u^{(k+1)} = \sigma\bigl(\sum_{u,v \trianglelefteq e}\Lambda_{uv}\F_{u \trianglelefteq e}^{\top}\F_{v \trianglelefteq e}\z_v^{(k)}\bigr)$. This is the blueprint with the message $\psi_e(a,b) = a$, so that $\mathbf{m}_{u \trianglelefteq e}^{(k)} = \F_{v \trianglelefteq e}\z_v^{(k)}$, the weights $\omega_{uv}^{(k)} = \Lambda_{uv}$, sum aggregation, and the update $\varphi_u(\x_u, \mathbf{m}_u) = \sigma(\mathbf{m}_u)$. When the attention coefficients are normalized by a softmax over the neighbourhood, the normalization is itself local: each edge also sends its unnormalized score, both quantities are summed, and the update divides one by the other, as in graph attention.

Res-SheafAN adds a residual connection and a skip rescaling to the SheafAN update, so that the layer computes:
\begin{equation*}
  \X^{(k+1)} = (1+\pmb{\varepsilon}^{(k)}) \odot \X^{(k)} + \sigma\parens*{\parens*{(\hat{\Lambda}^{(k)} \otimes \mathbf{1}_d\mathbf{1}_d^{\top}) \odot \rmA_{\gF^{(k)}} - \I}\Z^{(k)}}
\end{equation*}
so that $\x_u^{(k+1)} = (1+\pmb{\varepsilon}^{(k)}) \odot \x_u^{(k)} + \sigma\bigl(\sum_{u,v \trianglelefteq e}\Lambda_{uv}\F_{u \trianglelefteq e}^{\top}\F_{v \trianglelefteq e}\z_v^{(k)} - \z_u^{(k)}\bigr)$. The messages and the aggregation are those of SheafAN. The term $\z_u^{(k)} = \W_1^{(k)}\x_u^{(k)}\W_2^{(k)}$ depends only on the state of $u$, so it belongs to the node-local update $\varphi_u(\x_u, \mathbf{m}_u) = (1+\pmb{\varepsilon}^{(k)}) \odot \x_u + \sigma\bigl(\mathbf{m}_u - \W_1^{(k)}\x_u\W_2^{(k)}\bigr)$.
\end{proof}

\begin{proof}[Proof of Theorem~\ref{thm:recovery}, part (iii): Sheaf Adjacency.]
Sheaf adjacency propagation~\citep{bourgerie2026deep} replaces the Laplacian with the normalized sheaf adjacency $\hat{\rmA}_{\gF}$, whose $(u,v)$ block is $\rmD_u^{-1/2}\F_{u \trianglelefteq e}^{\top}\F_{v \trianglelefteq e}\rmD_v^{-1/2} = \hat{\F}_{u \trianglelefteq e}^{\top}\hat{\F}_{v \trianglelefteq e}$:
\begin{equation}
  \X^{(k+1)} = (1+\pmb{\varepsilon}^{(k)}) \odot \X^{(k)} + \alpha\,\sigma\parens*{\hat{\rmA}_{\gF^{(k)}}\Z^{(k)}}.
  \label{eqn:adj-layer}
\end{equation}
At node $u$ the argument of $\sigma$ is $\sum_{u,v \trianglelefteq e}\hat{\F}_{u \trianglelefteq e}^{\top}\hat{\F}_{v \trianglelefteq e}\z_v^{(k)}$. This is the blueprint with the normalized maps $\hat{\F}$, the message $\psi_e(a,b) = a$, sum aggregation, and the update $\varphi_u(\x_u, \mathbf{m}_u) = (1+\pmb{\varepsilon}^{(k)}) \odot \x_u + \alpha\,\sigma(\mathbf{m}_u)$.
\end{proof}

\begin{proof}[Proof of Theorem~\ref{thm:recovery}, part (iv): Neural Sheaf Propagation.]
Neural Sheaf Propagation (NSP)~\citep{suk2022surfing} follows the second-order recurrence: 
\begin{equation}
  \X^{(k+1)} = 2\X^{(k)} - \X^{(k-1)} - \sigma\parens*{\Delta_{\gF^{(k)}}\Z^{(k)}}, \qquad \X^{(-1)} = \mathbf{0}
  \label{eqn:nsp-layer}
\end{equation}
with $\Z^{(k)}$ as in Equation~\eqref{eqn:nsd-layer}. Because the update depends on $\X^{(k-1)}$, it is not directly of the form of Equation~\eqref{eqn:sheafmpnn-update}, where $\varphi_u$ receives only the current state and the aggregated message. We remove this dependence by augmenting the state.

Define $\tilde{\x}_u^{(k)} \coloneqq \bigl(\x_u^{(k)}, \x_u^{(k-1)}\bigr) \in \R^{2d \times f}$, with $\tilde{\x}_u^{(0)} = (\x_u^{(0)}, \mathbf{0})$, and let $\pi_1, \pi_2$ project onto the current and history blocks. The augmented sheaf has node stalks $\R^{2d}$, edge stalks $\R^{d}$, and restriction maps $\tilde{\F}_{u \trianglelefteq e} \coloneqq \hat{\F}_{u \trianglelefteq e} \circ \pi_1 : \R^{2d} \to \R^{d}$, which read the current block and apply the normalized original map. The transform uses $\tilde{\W}_1^{(k)} = \mathrm{diag}(\W_1^{(k)}, \mathbf{0})$, so that $\tilde{\z}_u^{(k)} = (\z_u^{(k)}, \mathbf{0})$. With the message $\psi_e(a,b) = b - a$ and sum aggregation:
\begin{equation*}
  \mathbf{m}_u^{(k+1)} = \sum_{u,v \trianglelefteq e}\tilde{\F}_{u \trianglelefteq e}^{\top}\bigl(\tilde{\F}_{u \trianglelefteq e}\tilde{\z}_u^{(k)} - \tilde{\F}_{v \trianglelefteq e}\tilde{\z}_v^{(k)}\bigr) = \bigl([\Delta_{\gF^{(k)}}\Z^{(k)}]_u,\; \mathbf{0}\bigr)
\end{equation*}
by Equation~\eqref{eq:normalized-local-laplacian}, since $\tilde{\F}_{u \trianglelefteq e}^{\top} = \pi_1^{\top}\hat{\F}_{u \trianglelefteq e}^{\top}$ embeds into the current block. The update then reads:
\begin{equation*}
  \varphi_u\bigl(\tilde{\x}_u, \mathbf{m}_u\bigr) = \Bigl(2\,\pi_1(\tilde{\x}_u) - \pi_2(\tilde{\x}_u) - \sigma\bigl(\pi_1(\mathbf{m}_u)\bigr),\; \pi_1(\tilde{\x}_u)\Bigr)
\end{equation*}
and it is a deterministic function of the augmented state and the aggregated message. By induction on $k$, $\tilde{\x}_u^{(k)} = (\x_u^{(k)}, \x_u^{(k-1)})$ for all $k$: the base case holds by the initialization, and if it holds at step $k$, then $\pi_1(\tilde{\x}_u^{(k+1)}) = 2\x_u^{(k)} - \x_u^{(k-1)} - \sigma([\Delta_{\gF^{(k)}}\Z^{(k)}]_u) = \x_u^{(k+1)}$ by Equation~\eqref{eqn:nsp-layer} and $\pi_2(\tilde{\x}_u^{(k+1)}) = \x_u^{(k)}$. The current block therefore reproduces NSP exactly.
\end{proof}

\begin{remark}
\label{rem:nsp-stalks}
The augmented restriction maps $\tilde{\F}_{u \trianglelefteq e} : \R^{2d} \to \R^{d}$ are not square, which the blueprint allows but which places NSP in a different sheaf regime from NSD, whose maps are endomorphisms. The degree of the augmented sheaf vanishes on the history block, so its own normalization is undefined; this is why the normalization of the original sheaf is folded into $\tilde{\F}$.
\end{remark}

\begin{remark}[Learnable Skip Rescaling].
\label{rem:eps-weighted-sheafmpnn}
Any update that keeps a residual connection to the current state can be extended with learnable per-dimension rescalings, since these are constants shared by all nodes and leave $\varphi_u$ a deterministic function of the state and the aggregated message. For NSP, the library's \texttt{add\_eps} option uses two rescalings $\pmb{\varepsilon}_1, \pmb{\varepsilon}_2 \in [-1,1]^d$, one for each block:
\begin{equation*}
  \pi_1(\tilde{\x}_u^{(k+1)}) = (2 + \pmb{\varepsilon}_1) \odot \pi_1(\tilde{\x}_u^{(k)}) - (1 + \pmb{\varepsilon}_2) \odot \pi_2(\tilde{\x}_u^{(k)}) - \sigma\bigl(\pi_1(\mathbf{m}_u^{(k+1)})\bigr)
\end{equation*}
which recovers Equation~\eqref{eqn:nsp-layer} at $\pmb{\varepsilon}_1 = \pmb{\varepsilon}_2 = \mathbf{0}$.
\end{remark}

\subsection{Polynomial Neural Sheaf Diffusion}
\label{app:polynsd}

Polynomial Neural Sheaf Diffusion (PolyNSD)~\citep{borgi2026polynomialneuralsheafdiffusion} replaces the single Laplacian application of NSD with a degree-$K$ polynomial filter, evaluated through the Chebyshev recurrence without eigendecomposition. We show that a PolyNSD layer is a sequence of $K$ SheafMPNN steps followed by a node-wise readout.

\paragraph{Setting.}
The spectrum of $\Delta_{\gF}$ lies in $[0, \lambda_{\max}]$ with $\lambda_{\max} = 2$ for the symmetric normalization. The rescaled Laplacian:
\begin{equation}
  \widehat{\Delta}_{\gF} \defeq \frac{2}{\lambda_{\max}}\,\Delta_{\gF} - \I_{nd}, \qquad \spec\bigl(\widehat{\Delta}_{\gF}\bigr) \subseteq [-1, 1]
  \label{eq:rescaled_lap}
\end{equation}
maps the spectrum onto the interval on which the Chebyshev polynomials $T_0(x) = 1$, $T_1(x) = x$, and $T_{k+1}(x) = 2x\,T_k(x) - T_{k-1}(x)$ satisfy $|T_k| \le 1$.

\paragraph{PolyNSD Update Rule.}
At layer $\ell$, let $\z_u = \W_1^{(\ell)}\x_u^{(\ell)}\W_2^{(\ell)}$ and define the Chebyshev iterates over the rescaled Laplacian:
\begin{equation}
  \bT_0 = \z, \qquad \bT_1 = \widehat{\Delta}_{\gF}\,\z, \qquad \bT_{k+1} = 2\,\widehat{\Delta}_{\gF}\,\bT_k - \bT_{k-1} \quad (k \ge 1).
  \label{eq:cheb_rec}
\end{equation}
The layer combines them with a high-pass skip term and a learnable convex combination:
\begin{equation}
  \mathbf{h}_u = \alpha_{\mathrm{hp}}\Bigl(\z_u - \tfrac{1}{\lambda_{\max}}[\Delta_{\gF}\z]_u\Bigr),
  \label{eq:hp_skip}
\end{equation}
\begin{equation}
  \x_u^{(\ell+1)} = \mathrm{diag}(\pmb{\gamma})\,\x_u^{(\ell)} - \sigma\Bigl(\sum_{k=0}^{K}\theta_k[\bT_k]_u + \mathbf{h}_u\Bigr),
  \label{eq:polynsd_update}
\end{equation}
where $\btheta = \mathrm{softmax}(\btheta_{\mathrm{raw}})$, so that $\theta_k \ge 0$ and $\sum_k \theta_k = 1$, $\pmb{\gamma} = 1 + \tanh(\pmb{\varepsilon})$ with $\pmb{\varepsilon} \in \R^{d}$ shared by all nodes, and $\alpha_{\mathrm{hp}} \in \R$ is learnable. Since $[\Delta_{\gF}\z]_u = \tfrac{\lambda_{\max}}{2}([\bT_1]_u + \z_u)$, the skip term equals $\mathbf{h}_u = \tfrac{\alpha_{\mathrm{hp}}}{2}([\bT_0]_u - [\bT_1]_u)$ for any $\lambda_{\max}$, a combination of the first two iterates.

\begin{definition}[Augmented PolyNSD State].
\label{def:poly_aug_state}
For $k \in \{1, \ldots, K\}$, the augmented state of node $u$ is:
\begin{equation}
  \tilde{\x}_u^{(k)} \defeq \bigl([\bT_k]_u,\; [\bT_{k-1}]_u,\; \mathbf{S}_u^{(k)}\bigr) \in \R^{3d \times f}, \qquad \mathbf{S}_u^{(k)} \defeq \sum_{j=0}^{k}\theta_j[\bT_j]_u + \mathbf{h}_u
\end{equation}
where the accumulator $\mathbf{S}_u^{(k)}$ already contains the skip term. The augmented sheaf $\tilde{\gF}$ has node stalks $\R^{3d}$, edge stalks $\R^{d}$, and restriction maps $\tilde{\F}_{u \trianglelefteq e} \defeq \hat{\F}_{u \trianglelefteq e} \circ \pi_1 : \R^{3d} \to \R^{d}$, where $\pi_1, \pi_2, \pi_3$ project onto the three blocks.
\end{definition}

\begin{theorem}[PolyNSD is a SheafMPNN].
\label{thm:polynsd}
A PolyNSD layer of degree $K \ge 1$ (Equation~\eqref{eq:polynsd_update}) equals $K$ SheafMPNN steps on the augmented sheaf $(\gG, \tilde{\gF})$, which compute $\tilde{\x}_u^{(K)}$ from $(\z_u, \mathbf{0}, \mathbf{0})$, followed by the node-wise readout $\x_u^{(\ell+1)} = \mathrm{diag}(\pmb{\gamma})\,\x_u^{(\ell)} - \sigma\bigl(\pi_3(\tilde{\x}_u^{(K)})\bigr)$. Every step uses identity weight matrices, the message $\psi_e(a,b) = b - a$, and sum aggregation; the steps differ only in their node-local updates.
\end{theorem}

\begin{proof}
\emph{Messages and Aggregation.} At every step, the first block of the state holds the current iterate $\bT$. Since $\tilde{\F}_{u \trianglelefteq e}$ reads only that block, Equation~\eqref{eq:normalized-local-laplacian} gives the aggregated message:
\begin{equation}
  \mathbf{m}_u = \sum_{u,v \trianglelefteq e}\tilde{\F}_{u \trianglelefteq e}^{\top}\bigl(\tilde{\F}_{u \trianglelefteq e}\tilde{\x}_u - \tilde{\F}_{v \trianglelefteq e}\tilde{\x}_v\bigr) = \bigl([\Delta_{\gF}\bT]_u,\; \mathbf{0},\; \mathbf{0}\bigr)
  \label{eq:poly_lap_expand}
\end{equation}
one message per incidence and one propagation per step. The node can then form $[\widehat{\Delta}_{\gF}\bT]_u = \tfrac{2}{\lambda_{\max}}\pi_1(\mathbf{m}_u) - \pi_1(\tilde{\x}_u)$ locally.

\emph{Updates.} The first step maps $(\z_u, \mathbf{0}, \mathbf{0})$ to $\tilde{\x}_u^{(1)} = \bigl([\bT_1]_u, \z_u, \theta_0\z_u + \theta_1[\bT_1]_u + \tfrac{\alpha_{\mathrm{hp}}}{2}(\z_u - [\bT_1]_u)\bigr)$ with $[\bT_1]_u = [\widehat{\Delta}_{\gF}\z]_u$. Each later step $k = 1, \ldots, K-1$ applies the Chebyshev recurrence to the first two blocks and adds $\theta_{k+1}[\bT_{k+1}]_u$ to the accumulator:
\begin{equation}
  \varphi_u\bigl(\tilde{\x}_u^{(k)}, \mathbf{m}_u\bigr) = \bigl(2[\widehat{\Delta}_{\gF}\bT_k]_u - [\bT_{k-1}]_u,\; [\bT_k]_u,\; \mathbf{S}_u^{(k)} + \theta_{k+1}\bigl(2[\widehat{\Delta}_{\gF}\bT_k]_u - [\bT_{k-1}]_u\bigr)\bigr).
  \label{eq:poly_update}
\end{equation}
Both updates are deterministic functions of the current augmented state and the aggregated message, as Equation~\eqref{eqn:sheafmpnn-update} requires.

\emph{Correctness.} By Equation~\eqref{eq:cheb_rec}, the first component of Equation~\eqref{eq:poly_update} is $\bT_{k+1}$, the second is $\bT_k$, and the third adds $\theta_{k+1}\bT_{k+1}$ to the accumulator. Since $\bT_k$ is generated by the Chebyshev recurrence in the matrix $\widehat{\Delta}_{\gF}$, induction gives $\bT_k = T_k(\widehat{\Delta}_{\gF})\,\z$, and after $K$ steps $\pi_3(\tilde{\x}_u^{(K)}) = \sum_{k=0}^{K}\theta_k[\bT_k]_u + \mathbf{h}_u$, the argument of $\sigma$ in Equation~\eqref{eq:polynsd_update}. The readout uses only $\x_u^{(\ell)}$ and the state of $u$, so it is node-wise and requires no further propagation.
\end{proof}

\begin{remark}[Reduction to NSD].
\label{rem:poly_to_nsd}
For $K = 1$ and $\lambda_{\max} = 2$, the argument of $\sigma$ is $(\theta_0 - \theta_1 + \alpha_{\mathrm{hp}})\,\z_u + (\theta_1 - \tfrac{\alpha_{\mathrm{hp}}}{2})[\Delta_{\gF}\z]_u$. Choosing $\alpha_{\mathrm{hp}} = \theta_1 - \theta_0$ cancels the first term and leaves $\tfrac{1}{2}[\Delta_{\gF}\z]_u$, because $\theta_0 + \theta_1 = 1$. Doubling $\W_2^{(\ell)}$ absorbs the factor $\tfrac{1}{2}$, so the layer computes $\mathrm{diag}(\pmb{\gamma})\,\x_u - \sigma([\Delta_{\gF}\z]_u)$: the NSD update of Equation~\eqref{eqn:nsd-layer} with unit step size, which is the original update of \citet{bodnar2022neural}. Every such NSD layer is therefore a degree-one PolyNSD layer.
\end{remark}

\begin{remark}[Restriction-map Variants].
\label{rem:poly_variants}
The proof uses only the linearity of the restriction maps, so Theorem~\ref{thm:polynsd} holds for all six NSD map variants (Table~\ref{tab:map-families}). For orthogonal maps $\F_{u \trianglelefteq e}^{\top}\F_{u \trianglelefteq e} = \I_d$, so $\rmD_u = \deg(u)\,\I_d$ and the normalization reduces to a scalar per node.
\end{remark}

\begin{remark}[Laplacian-free Filter Evaluation].
\label{rem:laplacian_free}
Each product $\widehat{\Delta}_{\gF}\bT_k$ is evaluated through Equation~\eqref{eq:poly_lap_expand}, which needs only the state of $u$, the states of its neighbours, and the maps on its incidences, so the $nd \times nd$ operator is never assembled. Per layer, the library stores the restriction-map products, of total size $O(|\gE|d^2)$, computed once by the map predictor and reused in all $K$ steps; the half-degree matrices $\rmD_u^{-1/2}$, of size $O(nd^2)$; and three stalk tensors of size $O(ndf)$, namely the two live iterates and the accumulator. A degree-$K$ layer therefore costs $O(K|\gE|d^2f)$ operations and $O(|\gE|d^2 + nd^2 + ndf)$ memory, excluding map prediction. By contrast, the original NSD implementation of \citet{bodnar2022neural} assembles a sparse $nd \times nd$ Laplacian with $O(nd^2 + |\gE|d^2)$ non-zeros and multiplies it with the features, and spectral filters based on eigendecomposition additionally pay $O((nd)^3)$ for the factorization. Both message passing and sparse multiplication perform the same $O(|\gE|d^2f)$ arithmetic per step; the difference lies in avoiding the assembly and storage of the global operator.
\end{remark}

\subsection{SheafPool as Graph-Level Sheaf Message Passing}
\label{app:sheafpool}

\textsc{SheafPool}~\citep{braithwaite2026heterogeneous} produces a graph-level representation from the node stalks of a sheaf backbone. We show that it is a single SheafMPNN layer in which every graph sends its aligned node stalks to a receive-only virtual node.
\paragraph{Setting.}
Let $\mathcal{B} = \{\gG_1, \ldots, \gG_B\}$ be a batch of graphs, $\gG_g = (\gV_g, \gE_g)$, and let $\mathbf{H}_i \in \R^{d \times f}$ be the stalk embedding of node $i$ produced by the backbone. The pooling graph $\gG^{+}$ adds one virtual node $\star_g$ per graph and the arcs:
\begin{equation}
  \gE^{+} = \bigl\{(i \to \star_g) : i \in \gV_g,\; g = 1, \ldots, B\bigr\}
  \label{eq:sheafpool-augmented-edges}
\end{equation}
and the pooling sheaf $\gF^{+}$ uses the stalk $\R^{d}$ at every node and edge, with identity restriction maps:
\begin{equation}
  \F_{i \trianglelefteq e} = \F_{\star_g \trianglelefteq e} = \I_d \qquad \text{for every arc } e = (i \to \star_g).
  \label{eq:sheafpool-trivial-restrictions}
\end{equation}
Messages are aggregated only along incoming arcs, as in the directed extension of Appendix~\ref{app:directed}, so the real nodes receive nothing and each virtual node is receive-only. Before pooling, \textsc{SheafPool} canonicalizes each node stalk. It whitens the stalk and aligns it to a learnable anchor $\mathbf{A} \in \R^{d \times f}$ by orthogonal Procrustes:
\begin{equation}
  \mathbf{Z}_i = \bigl(\mathbf{H}_i\mathbf{H}_i^{\top} + \varepsilon\I_d\bigr)^{-1/2}\mathbf{H}_i,
  \qquad
  \mathbf{R}_i^{\star} = \argmin_{\mathbf{R} \in \mathrm{O}(d)}\bigl\|\mathbf{R}^{\top}\mathbf{Z}_i - \mathbf{A}\bigr\|_F^2,
  \qquad
  \widetilde{\mathbf{Z}}_i = {\mathbf{R}_i^{\star}}^{\top}\mathbf{Z}_i
  \label{eq:sheafpool-whitening}
\end{equation}
and scores it through the channel energies $\mathbf{u}_i = \mathrm{diag}(\mathbf{Z}_i^{\top}\mathbf{Z}_i) \in \R^{f}$, which are invariant under orthogonal changes of stalk basis:
\begin{equation}
  \alpha_i = \frac{\exp(s_{\theta}(\mathbf{u}_i))}{\sum_{j \in \gV_g}\exp(s_{\theta}(\mathbf{u}_j))}, \qquad i \in \gV_g.
  \label{eq:sheafpool-invariant-attention}
\end{equation}
The pooled stalk, the update of a learnable graph token $\mathbf{T} \in \R^{d \times f}$, and the invariant readout are of the form:
\begin{align}
  \mathbf{P}_g &= \sum_{i \in \gV_g}\alpha_i\widetilde{\mathbf{Z}}_i,
  \label{eq:sheafpool-aggregation-map}\\
  \widehat{\mathbf{T}}_g &= \mathrm{LayerNorm}\bigl(\phi([\mathbf{T}, \mathbf{P}_g])\bigr),
  \label{eq:sheafpool-update-map}\\
  \mathbf{g}_g &= \rho(\widehat{\mathbf{T}}_g) = \mathrm{diag}\bigl(\widehat{\mathbf{T}}_g^{\top}\widehat{\mathbf{T}}_g\bigr) \in \R^{f}
  \label{eq:sheafpool-invariant-readout}
\end{align}
where $\phi$ is a channel-wise MLP. As \citet{braithwaite2026heterogeneous} stress, the token only receives information: it does not take part in the node-level diffusion and cannot act as a shortcut between nodes.

\begin{proposition}[SheafPool is a Graph-level SheafMPNN].
\label{prop:sheafpool}
\textsc{SheafPool} is a single SheafMPNN layer on the pooling graph $\gG^{+}$ with the sheaf of Equation~\eqref{eq:sheafpool-trivial-restrictions}, followed by the node-wise readout $\rho$ at the virtual nodes.
\end{proposition}

\begin{proof}
The canonicalization in Equation~\eqref{eq:sheafpool-whitening} and the scores $s_{\theta}(\mathbf{u}_i)$ depend only on $\mathbf{H}_i$, so they are node-local and can be computed by the update of the preceding layer; we take the pre-pooling state of node $i$ to be $(\widetilde{\mathbf{Z}}_i, s_{\theta}(\mathbf{u}_i))$. On each arc $(i \to \star_g)$, the identity map transports $\widetilde{\mathbf{Z}}_i$ unchanged, and the message function sends the pair $\bigl(\exp(s_{\theta}(\mathbf{u}_i))\,\widetilde{\mathbf{Z}}_i,\; \exp(s_{\theta}(\mathbf{u}_i))\bigr)$, which depends only on the source. Sum aggregation over the incoming arcs gives, at $\star_g$, the pair $\bigl(\sum_{i \in \gV_g}\exp(s_{\theta}(\mathbf{u}_i))\widetilde{\mathbf{Z}}_i,\; \sum_{i \in \gV_g}\exp(s_{\theta}(\mathbf{u}_i))\bigr)$. The update at $\star_g$ divides the first component by the second, which yields $\mathbf{P}_g$ of Equation~\eqref{eq:sheafpool-aggregation-map} with the softmax weights of Equation~\eqref{eq:sheafpool-invariant-attention}, and then applies Equation~\eqref{eq:sheafpool-update-map}. Real nodes have no incoming arcs and keep their states. Applying $\rho$ at each virtual node gives Equation~\eqref{eq:sheafpool-invariant-readout}.
\end{proof}

\begin{remark}
\label{rem:sheafpool-invariance}
The readout is permutation invariant when all node-wise choices are shared and deterministic. Invariance to general linear changes of stalk basis is not guaranteed by the regularized whitening: already for $d = f = 1$, $\mathbf{H} = h$ gives $h/\sqrt{h^2 + \varepsilon}$, whose magnitude changes under $h \mapsto ah$, and exact unregularized whitening requires full row rank. Orthogonal invariance additionally requires a unique Procrustes solution or an equivariant tie-breaking rule, which arbitrary SVD choices at degeneracies do not provide. Proposition~\ref{prop:sheafpool} therefore establishes locality, not unconditional gauge invariance. \ifdefined\preprintversion\else Injectivity over multisets of canonicalized node descriptors further requires an injective DeepSets-style multiset encoder~\citep{zaheer2017deepset}, mirroring the relationship between sum-based graph readouts and injective multiset functions~\citep{xu2019how}.\fi
\end{remark}

\subsection{Heterogeneous NSD and NSP as SheafMPNNs}
\label{app:hetsheaf}

Heterogeneous extensions of NSD and NSP~\citep{braithwaite2026heterogeneous} condition the restriction-map predictor on the type structure of the graph. Since the blueprint is parameterized by the restriction maps themselves and not by the mechanism that predicts them, any type-conditioned predictor yields a valid SheafMPNN.

\paragraph{Heterogeneous Graph Structure.}
Let $\gG = (\gV, \gE)$ be a graph with a node-type map $\tau : \gV \to \calT$ and an edge-relation map $\rho : \gE \to \calR$ into finite sets of types. Every node and edge carries the stalk $\R^{d}$.

\begin{definition}[Type-conditioned Restriction-map Predictor].
\label{def:het_predictor}
A type-conditioned predictor at layer $k$ produces, for each incidence $u \trianglelefteq e$ with $e = \{u, v\}$, a map $\F_{u \trianglelefteq e}^{(k)} = \Phi^{(k)}_{\rho(e)}\bigl(\mathbf{c}^{(k)}(u, v, e)\bigr) \in \R^{d \times d}$, where the context $\mathbf{c}^{(k)}$ combines any non-empty subset of the node features $[\x_u^{(k)} \,\|\, \x_v^{(k)}]$, the node-type one-hot encodings $[\mathbf{1}_{\tau(u)} \,\|\, \mathbf{1}_{\tau(v)}]$, and the relation one-hot encoding $\mathbf{1}_{\rho(e)}$. The predictor is either a single MLP shared by all edges, which may receive the relation encoding as input, or an ensemble $\{\Phi^{(k)}_r\}_{r \in \calR}$ with one MLP per relation type. Its output parameterizes a diagonal, general, attention-based, orthogonal, or low-rank map as in the homogeneous case.
\end{definition}

\noindent The type-conditioned Laplacian $L_{\gF_{\mathrm{het}}}$ and its normalization $\Delta_{\gF_{\mathrm{het}}}$ are defined as in the homogeneous case, with the type-conditioned maps in place of the original ones.

\paragraph{Het-NSD update rule.}
At layer $k$, Het-NSD computes the next state $\x_u^{(k+1)}$ from the current state $\x_u^{(k)}$ and the normalized Laplacian of the transformed state $\z_u^{(k)}$ as:
\begin{equation}
  \x_u^{(k+1)} = \x_u^{(k)} - \alpha\,\sigma\bigl([\Delta_{\gF_{\mathrm{het}}^{(k)}}\z^{(k)}]_u\bigr), \qquad \z_u^{(k)} = \W_1^{(k)}\x_u^{(k)}\W_2^{(k)}
  \label{eq:hetnsd_update}
\end{equation}
with a learnable step size $\alpha$ and restriction maps from a type-conditioned predictor.

\begin{theorem}[Het-NSD is a SheafMPNN].
\label{thm:hetnsd}
Het-NSD (Equation~\eqref{eq:hetnsd_update}) is a SheafMPNN on the type-conditioned sheaf $(\gG, \gF_{\mathrm{het}})$, with the stalk transform $\z_u^{(k)} = \W_1^{(k)}\x_u^{(k)}\W_2^{(k)}$, the normalized maps $\hat{\F}_{u \trianglelefteq e}^{(k)} = \F_{u \trianglelefteq e}^{(k)}\rmD_u^{-1/2}$, the message $\psi_e(a,b) = b - a$, sum aggregation, and the update $\varphi_u(\x_u, \mathbf{m}_u) = \x_u - \alpha\,\sigma(\mathbf{m}_u)$. Equivalently, after pullback, each edge contributes:
\begin{equation}
  \hat{\rmA}_{uu}^{(k)}\z_u^{(k)} - \hat{\rmA}_{uv}^{(k)}\z_v^{(k)}, \qquad
  \hat{\rmA}_{uu}^{(k)} = \hat{\F}_{u \trianglelefteq e}^{(k)\top}\hat{\F}_{u \trianglelefteq e}^{(k)}, \quad
  \hat{\rmA}_{uv}^{(k)} = \hat{\F}_{u \trianglelefteq e}^{(k)\top}\hat{\F}_{v \trianglelefteq e}^{(k)}
  \label{eq:hetnsd_msg}
\end{equation}
which is how the library evaluates it from precomputed products.
\end{theorem}

\begin{proof}
By Equation~\eqref{eq:normalized-local-laplacian} applied to the type-conditioned maps, we have:
\begin{equation}
  [\Delta_{\gF_{\mathrm{het}}^{(k)}}\z^{(k)}]_u = \sum_{u,v \trianglelefteq e}\hat{\F}_{u \trianglelefteq e}^{(k)\top}\bigl(\hat{\F}_{u \trianglelefteq e}^{(k)}\z_u^{(k)} - \hat{\F}_{v \trianglelefteq e}^{(k)}\z_v^{(k)}\bigr)
  \label{eq:hetnsd_lap_expand}
\end{equation}
which is the message $\psi_e(a,b) = b - a$ on the transported states, pulled back and summed; expanding the pullback gives Equation~\eqref{eq:hetnsd_msg}. The update $\x_u - \alpha\,\sigma(\mathbf{m}_u)$ is a deterministic function of the state and the aggregated message. The argument uses only that each $\F_{u \trianglelefteq e}^{(k)}$ is a linear map $\R^{d} \to \R^{d}$, which Definition~\ref{def:het_predictor} guarantees for every context and map family.
\end{proof}

\paragraph{Het-NSP Update Rule.}
Het-NSP keeps the second-order recurrence of NSP with type-conditioned maps:
\begin{equation}
  \x_u^{(k+1)} = 2\,\x_u^{(k)} - \x_u^{(k-1)} - \sigma\bigl([\Delta_{\gF_{\mathrm{het}}^{(k)}}\z^{(k)}]_u\bigr), \qquad \x_u^{(-1)} = \mathbf{0}
  \label{eq:hetnsp_update}
\end{equation}
and stores the augmented state $\tilde{\x}_u^{(k)} = (\x_u^{(k)}, \x_u^{(k-1)})$ as a single tensor of shape $[n, 2d, f]$.

\begin{definition}[Augmented Het-NSP Sheaf].
\label{def:hetnsp_aug_sheaf}
The augmented sheaf $\tilde{\gF}_{\mathrm{het}}$ has node stalks $\R^{2d}$, edge stalks $\R^{d}$, and restriction maps $\tilde{\F}_{u \trianglelefteq e}^{(k)} \defeq \hat{\F}_{u \trianglelefteq e}^{(k)} \circ \pi_1 : \R^{2d} \to \R^{d}$, where $\pi_1$ extracts the current block. The map predictor receives only the current block, so the history is never exposed to it.
\end{definition}

\begin{theorem}[Het-NSP is a SheafMPNN].
\label{thm:hetnsp}
Het-NSP (Equation~\eqref{eq:hetnsp_update}) is a SheafMPNN on the augmented sheaf $(\gG, \tilde{\gF}_{\mathrm{het}})$, with the transform $\tilde{\W}_1^{(k)} = \mathrm{diag}(\W_1^{(k)}, \mathbf{0})$, the message $\psi_e(a,b) = b - a$, sum aggregation, and the update:
\begin{equation}
  \varphi_u\bigl(\tilde{\x}_u, \mathbf{m}_u\bigr) = \Bigl(2\,\pi_1(\tilde{\x}_u) - \pi_2(\tilde{\x}_u) - \sigma\bigl(\pi_1(\mathbf{m}_u)\bigr),\; \pi_1(\tilde{\x}_u)\Bigr).
  \label{eq:hetnsp_update_fn}
\end{equation}
\end{theorem}

\begin{proof}
The proof of Theorem~\ref{thm:recovery}, part (iv), applies verbatim with the type-conditioned maps: the aggregated message is $\bigl([\Delta_{\gF_{\mathrm{het}}^{(k)}}\z^{(k)}]_u, \mathbf{0}\bigr)$ by Equation~\eqref{eq:hetnsd_lap_expand}, and the same induction shows that the current block of Equation~\eqref{eq:hetnsp_update_fn} reproduces Equation~\eqref{eq:hetnsp_update} while the second block stores the history. As for Het-NSD, only the linearity of the type-conditioned maps is used.
\end{proof}

\begin{remark}[Reduction to the Homogeneous Models].
\label{rem:het_to_homo}
With a single node type and a single relation type, the context reduces to the node features and the ensemble to one MLP, so the type-conditioned maps become the NSD maps of \citet{bodnar2022neural}. Het-NSD then coincides with NSD without skip rescaling ($\pmb{\varepsilon} = \mathbf{0}$ in Equation~\eqref{eqn:nsd-layer}), and Het-NSP with NSP.
\end{remark}

\begin{remark}[Type Conditioning preserves the Structure].
\label{rem:het_context_strategies}
All context strategies of Definition~\ref{def:het_predictor}, and their combinations, produce restriction maps that are deterministic functions of the current node features and the graph structure. The theorems therefore hold for every map family and for every combination of the library options \texttt{use\_features}, \texttt{use\_node\_types}, \texttt{use\_edge\_types}, and \texttt{ensemble}.
\end{remark}

\begin{remark}[Stalk Geometry of Het-NSP].
\label{rem:het_nsp_stalk_geometry}
As in Remark~\ref{rem:nsp-stalks}, the augmented maps are not square, and the degree of the augmented sheaf, $\sum_{u \trianglelefteq e}(\F_{u \trianglelefteq e}^{(k)\top}\F_{u \trianglelefteq e}^{(k)}) \oplus \mathbf{0} \in \R^{2d \times 2d}$, vanishes on the history block. Only the current block takes part in the Laplacian action, and the normalization is taken from the original type-conditioned sheaf.
\end{remark}

\subsection{Directed Extension}
\label{app:directed}
For a directed edge $e = (u \to v)$, the source and target incidences may carry different predictors, $\F_{u \trianglelefteq e}^{(k)} = \Phi_{\mathrm{src}}^{(k)}(\x_u \,\|\, \x_v)$ and $\F_{v \trianglelefteq e}^{(k)} = \Phi_{\mathrm{dst}}^{(k)}(\x_u \,\|\, \x_v)$, and both the predictors and the message function may depend on the order of the endpoints.

\begin{theorem}[Directed SheafMPNN].
\label{thm:dirsheafmpnn}
Any directed sheaf layer whose messages are computed independently on each arc and whose node update aggregates the incoming messages with a permutation-invariant operator is an instance of the SheafMPNN blueprint, including layers with asymmetric source and target predictors and an asymmetric message function.
\end{theorem}

\begin{proof}
For each arc $(v \to u)$, set $\mathbf{m}_{v \to u}^{(k)} = \psi_{v \to u}^{(k)}\bigl(\F_{v \trianglelefteq e}^{(k)}\z_v^{(k)}, \F_{u \trianglelefteq e}^{(k)}\z_u^{(k)}, \mathbf{e}_{uv}\bigr)$. Pulling it back through $\F_{u \trianglelefteq e}^{(k)\top}$ and aggregating over the incoming arcs of $u$ gives Equation~\eqref{eqn:sheafmpnn-aggregate}, and the node-local update gives Equation~\eqref{eqn:sheafmpnn-update}. Neither equation requires the maps, the messages, or the neighbourhood relation to be symmetric.
\end{proof}

\subsection{Cooperative Sheaf Neural Networks}
\label{app:cooperative}
Cooperative Sheaf Neural Networks (CSNNs)~\citep{ribeiro2026cooperative} use directed cellular sheaves to separate sending from receiving. On a directed graph, their out- and in-degree sheaf Laplacians act as:
\begin{align*}
  [L^{\mathrm{out}}_{\gF}\x]_u &= \sum_{v \in N^{+}(u)}\bigl(\F_{u \trianglelefteq uv}^{\top}\F_{u \trianglelefteq uv}\x_u - \F_{u \trianglelefteq vu}^{\top}\F_{v \trianglelefteq vu}\x_v\bigr), \\
  [L^{\mathrm{in}}_{\gF}\x]_u &= \sum_{v \in N^{-}(u)}\bigl(\F_{u \trianglelefteq vu}^{\top}\F_{u \trianglelefteq vu}\x_u - \F_{u \trianglelefteq uv}^{\top}\F_{v \trianglelefteq uv}\x_v\bigr)
\end{align*}
and learned gates select and combine these directed transports locally.

\begin{theorem}[Cooperative SNNs are SheafMPNNs].
\label{thm:cooperative}
A CSNN layer built from $L^{\mathrm{out}}_{\gF}$ and $L^{\mathrm{in}}_{\gF}$, with arbitrary node-local gates and update, is a directed SheafMPNN.
\end{theorem}

\begin{proof}
Each summand of either Laplacian is a message on one directed incidence: the difference between a self transport and a neighbour transport, pulled back to the receiving stalk. Each operator is therefore realized by the message and sum-aggregation stages of Theorem~\ref{thm:dirsheafmpnn}. A gate computed from the state of a node and its aggregated messages can enter the weights $\omega_{uv}$ or the update $\varphi_u$, and combining the two aggregated messages in that update is again a SheafMPNN layer. The independently parameterized in- and out-maps are what allow a node to control sending and receiving separately.
\end{proof}

\subsection{Joint Sheaf Neural Networks}
\label{app:jsnn}

Joint Sheaf Neural Networks (JSNNs)~\citep{caralt2024joint} infer the restriction maps through a joint diffusion process inspired by opinion dynamics, rather than through an unrestricted map learner. We model the map-generating process abstractly by an edge state $\mathbf{s}_e$, updated as $\mathbf{s}_e^{(k+1)} = g_e^{(k)}(\mathbf{s}_e^{(k)}, \x_u^{(k)}, \x_v^{(k)})$, from which the maps are read as $\F_{u \trianglelefteq e}^{(k)} = \Phi_{u \trianglelefteq e}^{(k)}(\mathbf{s}_e^{(k)})$.

\begin{theorem}[JSNNs are SheafMPNNs].
\label{thm:jsnn}
Every JSNN layer whose joint process yields linear incidence maps and whose feature update is sheaf diffusion is a SheafMPNN; if the joint process has memory, the result holds on the augmented state $(\x_u, \mathbf{s}_e)$.
\end{theorem}

\begin{proof}
At a fixed layer, the state $\mathbf{s}_e^{(k)}$ determines linear maps $\F_{u \trianglelefteq e}^{(k)}$, and the feature update is the local transport, difference, pullback, and sum of part (i) of Theorem~\ref{thm:recovery}. The state update is edge-local and only determines the maps of the next layer, so it does not change this computation. When the state is recurrent, keeping $\mathbf{s}_e$ as an auxiliary edge state gives an augmented local transition, as for NSP.
\end{proof}

\subsection{Bundle Neural Networks}
\label{app:bunn}

Bundle Neural Networks (BuNNs)~\citep{bamberger2025bundle} diffuse features on a flat orthogonal vector bundle. In a local trivialization, each node carries an orthogonal frame $\mathbf{Q}_u \in \mathrm{O}(d)$, and parallel transport along $\{u, v\}$ is $\mathbf{P}_{u \leftarrow v} = \mathbf{Q}_u^{\top}\mathbf{Q}_v$, which is path-independent by flatness.

\begin{theorem}[BuNN Message Diffusion is a SheafMPNN].
\label{thm:bunn}
When the heat kernel is evaluated by a truncated Taylor series, a BuNN layer is a sequence of SheafMPNN steps on the flat orthogonal sheaf with restriction-map products $\F_{u \trianglelefteq e}^{\top}\F_{v \trianglelefteq e} = \mathbf{P}_{u \leftarrow v}$, followed by a node-wise update.
\end{theorem}

\begin{proof}
Choose the restriction maps in the frame of each edge so that their products equal $\mathbf{P}_{u \leftarrow v}$; for instance, $\F_{u \trianglelefteq e} = \mathbf{Q}_u$. The bundle Laplacian then acts at $u$ as $\sum_{v \sim u}(\z_u - \mathbf{P}_{u \leftarrow v}\z_v)$, which is the coboundary message of Equation~\eqref{eqn:sheafmpnn-message}, pulled back and summed as in Equation~\eqref{eqn:sheafmpnn-aggregate}. The truncated Taylor series of the heat kernel is a polynomial in this Laplacian, so it is evaluated by repeated propagation with a running accumulator, exactly as in Theorem~\ref{thm:polynsd}. Orthogonality and flatness restrict the map family and enable the factorization of BuNNs, but do not change the local form of the computation. The exact spectral evaluation of the heat kernel instead eigendecomposes the graph Laplacian and is not a local computation.
\end{proof}

\subsection{Bayesian Sheaf Neural Networks}
\label{app:bayesian}

Bayesian Sheaf Neural Networks (BSNNs)~\citep{gillespie2024bayesian} treat the sheaf as a latent random variable with a learned variational distribution $q_{\phi}(\gF \mid \X, \mathbf{y})$. At each layer, a sampled sheaf $\gF^{(k)} \sim q_{\phi}$ supplies the maps of an otherwise standard sheaf diffusion layer.

\begin{theorem}[BSNNs are Conditionally SheafMPNNs].
\label{thm:bayesian}
Conditional on every draw of its latent sheaf, a BSNN layer is a SheafMPNN, and its posterior-predictive output is the Monte Carlo average of these SheafMPNN evaluations.
\end{theorem}

\begin{proof}
Fix a draw $\gF^{(k)}$. Its restriction maps are ordinary linear maps on the incidences, so part (i) of Theorem~\ref{thm:recovery} applies verbatim. Reparameterized sampling changes how the maps are learned and how gradients are estimated, not the local computation once the maps are drawn. For $S$ draws, the posterior-predictive estimate $S^{-1}\sum_{s=1}^{S}f_{\gF_s}(\X)$ averages the conditional SheafMPNN outputs, which amounts to a final permutation-invariant aggregation over samples.
\end{proof}

\section{Library Design and Usage: Additional Details}
\label{app:library}

This appendix complements Section~\ref{sec:design}. \libname{} separates three concerns: predicting the restriction maps, propagating stalk-valued states by local message passing, and training models on specific tasks. We describe the package layout, the implemented model families, the anatomy of an NSD layer in terms of the blueprint of Equations~\eqref{eqn:sheafmpnn-transform}--\eqref{eqn:sheafmpnn-update}, the end-to-end models, the experiment tooling, and the development workflow.

\subsection{Package Layout}
\label{app:library-layout}

The source tree contains three packages (Table~\ref{tab:package-layout}): \texttt{sheaf\_mpnn} implements the homogeneous models and their convolutions, \texttt{hetsheaf\_mpnn} their type-conditioned counterparts, and \texttt{exp} the data loading, model construction, training, and evaluation. Because the experiment code is kept apart, a convolution can be used in any PyTorch Geometric model without adopting the supplied training loop.

\noindent Each architecture family pairs a layer module, which defines propagation on stalk-valued states, with a model module, which adds the feature encoder and the prediction head. Most homogeneous convolutions derive from \texttt{BaseSheafConv}, itself a subclass of PyTorch Geometric's \texttt{MessagePassing}. Directed convolutions derive from \texttt{BaseDirectedSheafConv}, and heterogeneous layers add type-aware restriction predictors. The base classes share the stalk transforms and common utilities, but leave each family free to define its own propagation rule.

\begin{table}[ht]
\centering\small
\caption{Sheaf-MPNN Main Modules.}
\vspace{0.2cm}
\label{tab:package-layout}
\begin{tabular}{@{}>{\raggedright\arraybackslash}p{0.34\textwidth}>{\raggedright\arraybackslash}p{0.60\textwidth}@{}}
\toprule
\textbf{Module} & \textbf{Responsibility} \\
\midrule
\texttt{sheaf\_mpnn.base\_conv} & Shared stalk transforms, map-predictor inputs, and propagation interfaces. \\
\texttt{sheaf\_mpnn.<family>} & Architecture-specific convolutions and end-to-end models. \\
\texttt{sheaf\_mpnn.utils} & Normalization, orthogonal parameterizations, connection maps, and training utilities. \\
\texttt{sheaf\_mpnn.sheaf\_pool} & Graph-level pooling of stalk-valued representations. \\
\texttt{hetsheaf\_mpnn} & Type-conditioned layers and model wrappers. \\
\texttt{exp.config}, \texttt{exp.registries} & Typed experiment settings and factories for models, datasets, and presets. \\
\texttt{exp.run}, \texttt{exp.het\_run}, \texttt{exp.gc\_run} & Runners for homogeneous node classification, heterogeneous node classification, and graph classification. \\
\texttt{exp.sweeps}, \texttt{exp.benchmark} & Hyperparameter search and benchmarks against reference implementations. \\
\bottomrule
\end{tabular}
\end{table}

\subsection{Model Families and Restriction Maps}
\label{app:library-families}

The model registry contains 15 homogeneous families, each with its heterogeneous counterpart, for 29 selectable model types (Table~\ref{tab:families}). Following \citet{braithwaite2026heterogeneous}, the heterogeneous models condition their restriction-map predictors on type information (Appendix~\ref{app:hetsheaf}). Their registry keys prefix the homogeneous name with \texttt{het\_}, except for polynomial diffusion, whose key is \texttt{poly\_het\_nsd}.

\noindent NSD, PolyNSD, Deep NSD, NSP, sheaf adjacency, and directed NSD share the six restriction-map variants of NSD: diagonal, general, orthogonal, low-rank, general-attention, and orthogonal-attention (Appendix~\ref{app:library-layer}). The two attention variants change only how the maps are generated and are unrelated to the SheafAN and Res-SheafAN architectures. Bayesian and nonlinear diffusion support the four variants without attention; for nonlinear diffusion, the general and low-rank maps extend the diagonal and orthogonal maps of the reference model. The remaining families define their own parameterizations, such as the fixed maps of connection-Laplacian NSD, the flat bundle transports of BuNN, and the joint update of joint diffusion.

\begin{table}[ht]
\centering\small
\caption{\emph{Homogeneous Architecture Families.} The table lists the 15 homogeneous families and their corresponding model types and restriction-map options. In the last column, ``six'' denotes the six NSD restriction-map variants of Table~\ref{tab:map-families}, and ``four'' the diagonal, general, orthogonal, and low-rank maps.}
\label{tab:families}
\begin{tabular}{@{}>{\raggedright\arraybackslash}p{0.35\textwidth}>{\raggedright\arraybackslash}p{0.22\textwidth}>{\raggedright\arraybackslash}p{0.30\textwidth}@{}}
\toprule
\textbf{Family} & \textbf{Model type} & \textbf{Map and model options} \\
\midrule
Neural Sheaf Diffusion & \texttt{nsd} & Six \\
Polynomial NSD & \texttt{poly\_nsd} & Six \\
Connection-Laplacian NSD & \texttt{conn\_nsd} & Precomputed maps \\
Deep NSD & \texttt{dnsd} & Six \\
Neural Sheaf Propagation & \texttt{nsp} & Six \\
Sheaf adjacency & \texttt{sheaf\_adj} & Six \\
Sheaf attention & \texttt{sheaf\_attn} & SheafAN, Res-SheafAN \\
Bayesian sheaf networks & \texttt{bsnn} & Four \\
Nonlinear sheaf diffusion & \texttt{nlsd} & Four \\
Bundle neural networks & \texttt{bunn} & Taylor or spectral solver \\
Directed NSD & \texttt{directed\_nsd} & Six \\
Cooperative sheaf networks & \texttt{csnn} & Conformal or non-conformal \\
Identity sheaf & \texttt{id\_sheaf} & Identity restrictions \\
Joint diffusion & \texttt{joint\_diff} & Joint update \\
Rotation-invariant sheaf & \texttt{rot\_inv} & Rotation-invariant update \\
\bottomrule
\end{tabular}
\end{table}

\noindent The directed families modify the propagation rule as well as the map predictor (Appendix~\ref{app:directed}). Also, every family propagates information through local messages and never assembles a global sheaf Laplacian. Only the optional spectral solver of BuNN departs from this, since it computes eigenpairs of the graph Laplacian with a full or truncated eigendecomposition, depending on the graph size and the configuration. The default BuNN solver stays local and evaluates a truncated Taylor expansion of the heat kernel instead (Appendix~\ref{app:bunn}).

\subsection{Anatomy of an NSD Layer}
\label{app:library-layer}

An NSD convolution is called as \texttt{layer(x\_feat, x\_stalk, edge\_index)}. The tensor \texttt{x\_stalk} holds the node states $\x_u\in\R^{d\times f}$ and has shape $[n,d,f]$, while \texttt{x\_feat} holds the context vectors from which the restriction maps are predicted and has shape $[n,c]$, where $c$ is \texttt{context\_dim}. The end-to-end model flattens the current state, so that $c=df$, but any context can be supplied. When the fixed channels of Table~\ref{tab:model-options} are enabled, the stalk dimension grows to $d_{\mathrm{tot}}=d+\mathbf{1}_{\mathrm{add\_lp}}+\mathbf{1}_{\mathrm{add\_hp}}$, which replaces $d$ in all tensor shapes.

\noindent Following PyTorch Geometric, an entry $(v,u)$ of \texttt{edge\_index} sends a message from the source $v$ to the receiver $u$, so an undirected graph must list both orientations of every edge. The predictor evaluates the two endpoint contexts in both orders to obtain the two restriction maps of each edge. It is a bias-free linear layer followed by $\tanh$, except in the attention variants, which pass its output through a softmax-based construction (Table~\ref{tab:map-families}).

\noindent The forward pass first applies the transform of Equation~\eqref{eqn:sheafmpnn-transform}, $\z_u=\W_1\,\mathrm{drop}(\x_u)\,\W_2$, where dropout acts on the diffusion input but not on the residual state; at initialization, $\W_1$ is the identity and $\W_2$ a random orthogonal matrix. Instead of applying the restriction maps separately in every message, the layer precomputes the per-edge products:
\[
 S_{ue}=\F_{u\trianglelefteq e}^{\top}\F_{u\trianglelefteq e},
 \qquad C_{uve}=\F_{u\trianglelefteq e}^{\top}\F_{v\trianglelefteq e}.
\]
With the block degree $D_u=\sum_{e\ni u}S_{ue}$ and $N_u=(D_u+I)^{-1/2}$, the pulled-back message of Equations~\eqref{eqn:sheafmpnn-message}--\eqref{eqn:sheafmpnn-aggregate}, with $\psi_e(a,b)=b-a$ and normalized maps $\F_{u\trianglelefteq e}N_u$, becomes:
\begin{equation}
 \mathbf{m}_{v\to u}=N_uS_{ue}N_u\z_u-N_uC_{uve}N_v\z_v.
 \label{eq:library-nsd-message}
\end{equation}
Summing these messages at $u$ yields the action of the augmented normalized sheaf Laplacian $NL_{\gF}N$ without forming the $nd\times nd$ operator, and the identity term keeps $N_u$ well defined even when the restriction maps are singular. Two numerical safeguards, adapted from the reference implementation of \citet{bodnar2022neural}, modify these products: for general and low-rank maps, the identity term receives a small jitter during training, and the normalized products of general maps are clipped entrywise to $[-1,1]$. Equation~\eqref{eq:library-nsd-message} is therefore exact at evaluation time whenever the clipping is inactive.

\noindent With the default residual connection, the layer then instantiates Equation~\eqref{eqn:sheafmpnn-update} as:
\begin{equation}
 \x_u^{\mathrm{new}}=
 \bigl(1+\tanh(\boldsymbol{\varepsilon})\bigr)\odot\x_u
 -\alpha\,\mathrm{ELU}\!\left(\sum_{v\to u}\mathbf{m}_{v\to u}\right)
\end{equation}
where $\boldsymbol{\varepsilon}\in\R^{d}$ is a learnable skip scaling, initialized at zero and broadcast across channels, and the step size $\alpha$ is learned by default in \texttt{NSDModel}. When the residual is disabled, the layer returns only the diffusion term $-\alpha\,\mathrm{ELU}(\cdot)$, so that the caller can supply its own update.

\begin{table}[ht]
\centering\small
\caption{\emph{NSD Restriction-map Variants}. The last column gives the number of predictor outputs per restriction map, excluding fixed channels and edge weights. Low-rank maps need fewer outputs, but their products $S_{ue}$ and $C_{uve}$ are still stored as dense $d\times d$ blocks.}
\label{tab:map-families}
\begin{tabular}{@{}>{\raggedright\arraybackslash}p{0.29\textwidth}>{\raggedright\arraybackslash}p{0.51\textwidth}>{\raggedright\arraybackslash}p{0.13\textwidth}@{}}
\toprule
\textbf{Variant} & \textbf{Restriction parameterization} & \textbf{Outputs} \\
\midrule
\texttt{DIAGONAL} & $\operatorname{diag}(w)$, with $w\in(-1,1)^d$ & $d$ \\
\texttt{GENERAL} & Dense matrix with entries in $(-1,1)$ & $d^2$ \\
\texttt{GENERAL\_ATTENTION} & $I-\operatorname{softmax}(P)$, with row-wise softmax & $d^2$ \\
\texttt{ORTHOGONAL} & Cayley or Householder construction & $\binom{d}{2}$ or $d^2$ \\
\texttt{ORTHOGONAL\_ATTENTION} & Cayley transform of the skew part of $I-\operatorname{softmax}(P)$ & $d^2$ \\
\texttt{LOW\_RANK} & $AB^{\top}$, with $A,B\in\R^{d\times r}$ & $2dr$ \\
\bottomrule
\end{tabular}
\end{table}

\noindent The orthogonal class offers four strategies through \texttt{orth\_strategy}. The \texttt{cayley} strategy uses a clamped half-Cayley transform, and \texttt{fasth} composes Householder reflections given by $d$ free reflector vectors. The \texttt{fasthpp} and \texttt{householder} strategies use compact reflector parameters; \texttt{householder} calls \texttt{torch\_householder} when it is installed and otherwise falls back to a native implementation. For unweighted orthogonal maps, $S_{ue}=I$ and the normalization reduces to $(\deg(u)+1)^{-1/2}I$; learned edge weights replace $\deg(u)$ by a weighted degree. Connection-Laplacian NSD uses a separate convolution whose orthogonal transports are precomputed from local geometric estimates rather than learned.

\subsection{End-to-End Models and Direct Layer Use}
\label{app:library-model}

\texttt{NSDModel} applies input dropout, a linear encoder, an ELU, and dropout, and then reshapes the representation into stalks; an optional second encoder layer acts before the reshape. Each diffusion layer receives the current stalk state together with its flattened context, and from the second layer onwards the context also receives dropout. A linear decoder maps the final flattened state to the predictions, or the concatenation of all layer outputs when jumping knowledge is enabled. Table~\ref{tab:model-options} lists the remaining options. The other families keep their own states and update rules, and hence their own options.

\begin{table}[ht]
\centering\small
\caption{Options of \texttt{NSDModel} beyond the core dimensions \texttt{in\_channels}, \texttt{out\_channels}, \texttt{stalk\_dim}, \texttt{hidden\_dim}, and \texttt{num\_layers}.}
\label{tab:model-options}
\begin{tabular}{@{}>{\raggedright\arraybackslash}p{0.31\textwidth}>{\raggedright\arraybackslash}p{0.63\textwidth}@{}}
\toprule
\textbf{Option} & \textbf{Effect} \\
\midrule
\texttt{variant}, \texttt{rank}, \texttt{orth\_strategy} & Select the restriction-map variant and its parameterization. \\
\texttt{alpha}, \texttt{learn\_alpha}, \texttt{add\_eps} & Set the diffusion step and enable the learnable skip scaling. \\
\texttt{use\_edge\_weights} & Enable learned symmetric edge weights for orthogonal maps. \\
\texttt{add\_lp}, \texttt{add\_hp} & Append a fixed channel with signless ($D+A$) or standard ($D-A$) graph-Laplacian blocks, respectively. \\
\texttt{sparse\_learner} & Sum the paired contexts over the stalk axis, reducing the predictor width from $2d_{\mathrm{tot}}f$ to $2f$ for flattened stalk inputs. \\
\texttt{second\_linear} & Add a second encoder layer. \\
\texttt{input\_dropout}, \texttt{dropout} & Set the input and internal dropout rates. \\
\texttt{normalize\_output}, \texttt{jknet} & Normalize the output representations, and concatenate the representations of all diffusion layers. \\
\texttt{add\_self\_loops} & Insert self-loops; off by default, since $N_u$ already includes the identity term. \\
\bottomrule
\end{tabular}
\end{table}

\noindent Convolutions can also be used on their own. In the following example, the caller supplies the encoder and the context, and no fixed channels are used:
\begin{lstlisting}[style=sheafpython]
from torch import nn
from sheaf_mpnn.nsd import GeneralNSDConv

d, f = 4, 16
encoder = nn.Linear(x.size(-1), d * f)
layer = GeneralNSDConv(stalk_dim=d, in_channels=f,
                       hidden_dim=f, context_dim=d * f)
x_stalk = encoder(x).reshape(-1, d, f)
x_feat = x_stalk.reshape(x_stalk.size(0), d * f)
x_stalk = layer(x_feat, x_stalk, edge_index)
\end{lstlisting}
Here \texttt{x} and \texttt{edge\_index} are the node features and edges of the input graph. The \texttt{hidden\_dim} argument is kept for interface compatibility and adds no hidden layer to the linear restriction predictor of NSD.

\subsection{Experiment Tooling}
\label{app:library-exp}

The \texttt{sheaf} command provides four tools: \texttt{run}, \texttt{splits}, \texttt{sweep}, and \texttt{benchmark}. \texttt{sheaf run} trains homogeneous node-classification models, while heterogeneous node classification and graph classification have their own entry points, \texttt{exp.het\_run} and \texttt{exp.gc\_run}, which share the same configuration components:
\begin{lstlisting}[style=sheafpython, language=bash]
sheaf run --preset texas --model.variant orthogonal --model.stalk-dim 5
sheaf splits --datasets cora texas
sheaf sweep --yaml-path configs/cora/nsd/general.yaml --preset cora_nsd_general
python -m exp.het_run --help
python -m exp.gc_run --help
sheaf benchmark report --manifest configs/benchmark/nsd_vs_original.yaml
\end{lstlisting}

\noindent Typed configurations describe the data, model, training procedure, hardware, and logging, and presets supply defaults that can be overridden from the command line. Training uses Adam, with a separate weight decay for the restriction-map predictors, and early stopping on a validation criterion. Results are summarized over the folds or seeds of each dataset's split protocol; for example, the OGB node datasets have a single official split, whereas \textsc{Pokec} has five.

\noindent The homogeneous runner covers the citation and heterophily benchmarks, the filtered \textsc{Chameleon} and \textsc{Squirrel} graphs, the five datasets of \citet{platonov2023critical}, the OGB node-property datasets, \textsc{Pokec}, \textsc{Telegram}, and a grid of synthetic directed graphs. Edges are symmetrized by default, and the \texttt{--directed} flag preserves the loaded orientation where one is available, so an experiment must state which of the two settings it uses. The heterogeneous runner supports the HGB datasets \textsc{IMDB}, \textsc{DBLP}, \textsc{ACM}, and \textsc{Freebase}, and the graph-classification runner supports the TU datasets with mean, sum, max, SAG, and \textsc{SheafPool} readouts. Data modules and a training module for link prediction are also included, although no runner exposes them yet.

\noindent Search spaces are stored as YAML files under \texttt{configs/}. The sweep tool uses the TPE sampler and median pruner of Optuna, supports shared study storage and optional Weights \& Biases logging, and re-evaluates the best configuration over ten seeds or folds. Each file fixes its own search space and number of trials, so comparisons across models should report both.

\noindent The benchmark manifest \texttt{configs/benchmark/nsd\_vs\_original.yaml} compares diagonal, orthogonal, and general NSD with the original implementation of \citet{bodnar2022neural}, which is included as a git submodule. The benchmark tool submits the runs as SLURM jobs, collects their outputs, and generates tables and plots of convergence, epoch time, peak GPU memory, and inference latency.

\subsection{Development and Reproducibility}
\label{app:library-dev}

The environment is managed with \texttt{uv} and pins Python~3.13, PyTorch~2.8.0, and Lightning~2.6.0, with CUDA~12.8 wheels on Linux and Windows and CPU wheels on macOS. Linting and formatting use \texttt{ruff}, and type checking uses \texttt{ty}. GitHub Actions runs the unit tests with coverage on pushes and pull requests, the integration tests daily, and the benchmark integration tests weekly.

\noindent Ten tutorial notebooks accompany the code, from the basics of GNNs to the principal sheaf model families; two of them cover hypergraph and directed hypergraph sheaves, which are not yet registered model types. Reproducing an experiment requires the code version, environment, dataset and preprocessing, model and map variant, splits and seeds, tuning budget, and checkpoint-selection rule. Numerical comparisons with a reference implementation additionally require matched weights, normalization, random choices, and error tolerances. As noted in Section~\ref{sec:design}, this paper reports neither test coverage nor numerical parity with reference implementations.

\subsection*{Acknowledgements}
\phantomsection\addcontentsline{toc}{subsection}{Acknowledgements}
\label{app:library-acks}

We thank Gabriele Onorato, Dario Loi, Ferran Hernandez Caralt, and James Rowbottom for their help in developing \libname{}. Their contributions range from implementing and testing model families to preparing experiment configurations and maintaining the continuous-integration workflows, and we are grateful for their feedback and discussions throughout the project.

\ifdefined\preprintversion\else\input{appendix_B_expressivity}\fi
\ifdefined\preprintversion\else\input{appendix_C_experiments}\fi

\vskip 0.2in
\bibliography{references}

\end{document}